%% file: paper.tex
\documentclass[]{fairmeta}
\usepackage{wrapfig}
\usepackage{tabularx}
\usepackage{textcomp}
\usepackage{stfloats}
\usepackage{url}
\usepackage{verbatim}
\usepackage{titlesec}
\usepackage{tocloft}
\usepackage{adjustbox}
\usepackage{multirow}
\usepackage{pifont}
\usepackage{tikz}
\usepackage{comment}
\usepackage{amsmath,amssymb} % define this before the line numbering.
\usepackage{colortbl}  %彩色表格需要加载的宏包
\usepackage{color}
\usepackage{booktabs} 
\usepackage{natbib} 
\setcitestyle{square,comma,numbers,sort&compress}
\usepackage{graphicx}    % 用于插入图片
\usepackage{subcaption} 
\usepackage{multirow} % 表格多行合并
\usepackage{booktabs} % 表格更好的线条
\usepackage{subcaption} % 支持 subtable 和 subfigure
\RequirePackage{xspace}
\makeatletter
\DeclareRobustCommand\onedot{\futurelet\@let@token\@onedot}
\def\@onedot{\ifx\@let@token.\else.\null\fi\xspace}

\def\ie{\emph{i.e}\onedot}

\makeatother

\definecolor{adptorange}{RGB}{248, 205, 172}
\definecolor{cmpblue}{RGB}{189, 215, 238}
\definecolor{cmpblue}{RGB}{189, 215, 238}

\definecolor{our_red}{RGB}{232,157,160}
\definecolor{our_blue}{RGB}{136,206,230}
\definecolor{our_orange}{RGB}{246,200,168}
\definecolor{our_green}{RGB}{178,211,164}

\definecolor{attn_code0}{RGB}{247,215,200}
\definecolor{attn_code1}{RGB}{238,169,139}
\definecolor{mlp_code0}{RGB}{204,201,221}
\definecolor{mlp_code1}{RGB}{102,95,153}

\definecolor{token_blue}{RGB}{84, 120, 140}
\input{math_commands}
\input{requirements/table}

\usepackage{pifont}
\newcommand{\one}{\textcolor{violet}{\ding{182}}}
\newcommand{\two}{\textcolor{violet}{\ding{183}}}
\newcommand{\three}{\textcolor{violet}{\ding{184}}}

\newcommand{\listnumber}[1]{\textbf{\color{violet}{#1}}}

\newcommand{\np}{\phantom{.}}

\newcommand{\nt}{\phantom{--}}

\usepackage{pifont}       % \ding{xx}
\usepackage{bbding}       % \Checkmark,\XSolid,... (需要和pifont宏包共同使用)
\usepackage{fontawesome}
\usepackage{xspace}

\usepackage{float}

\newlength\savewidth

\newcolumntype{x}[1]{>{\centering\arraybackslash}p{#1pt}}
\newcolumntype{y}[1]{>{\raggedright\arraybackslash}p{#1pt}}
\newcolumntype{z}[1]{>{\raggedleft\arraybackslash}p{#1pt}}

\renewcommand{\paragraph}[1]{\vspace{1mm}\noindent\textbf{#1}}
\usepackage{colortbl}
\usepackage{xcolor}
\usepackage{wrapfig}

\renewcommand{\paragraph}[1]{\vspace{1.25mm}\noindent\textbf{#1}}

\usepackage{algorithm}
\usepackage{listings}

\definecolor{codeblue}{rgb}{0.25, 0.5, 0.5}
\definecolor{codekw}{rgb}{0.35, 0.35, 0.75}
\lstdefinestyle{Pytorch}{
    language = Python,
    backgroundcolor = \color{white},
    basicstyle = \fontsize{9pt}{8pt}\selectfont\ttfamily\bfseries,
    columns = fullflexible,
    aboveskip=1pt,
    belowskip=1pt,
    breaklines = true,
    captionpos = b,
    commentstyle = \color{codeblue},
    keywordstyle = \color{codekw},
}

\definecolor{green}{HTML}{009000}
\definecolor{red}{HTML}{ea4335}

\definecolor{cvblue}{rgb}{0.15, 0.45, 0.68}
\title{VideoLLaMA 3: Frontier Multimodal Foundation Models for Image and Video Understanding}

\author[*]{Boqiang Zhang}
\author[*]{Kehan Li}
\author[*]{Zesen Cheng}
\author[*]{Zhiqiang Hu}
\author[*]{Yuqian Yuan}
\author[*]{Guanzheng Chen}
\author[*]{Sicong Leng}
\author[*, \dagger]{Yuming Jiang}
\author[*, \dagger]{Hang Zhang}
\author[*, \dagger]{Xin Li}
\author{Peng Jin}
\author{Wenqi Zhang}
\author{Fan Wang}
\author{Lidong Bing}
\author{Deli Zhao}

\affiliation[]{DAMO Academy, Alibaba Group\\}
\affiliation[]{Hupan Lab}
\contribution[*]{Equal contribution}
\contribution[\dagger]{Proj Lead}

\input{sections/abstract}

\date{\today}

\begin{document}
\thispagestyle{firstheader}
\maketitle
\pagestyle{empty}

\input{sections/introduction}

\input{sections/method}
\input{sections/data_mixture}

\input{sections/experiment}

\input{sections/related_work}
\input{sections/conclusion}

\bibliographystyle{unsrt}
\bibliography{paper}

% \printbibliography[heading=bibintoc]

% % \clearpage
% \newpage
% \beginappendix
% \input{sections/appendix}

\end{document}

%% file: math_commands.tex
\usepackage{amsmath,amsfonts,bm}

\def\eqref#1{equation~\ref{#1}}
\def\1{\bm{1}}

\DeclareMathAlphabet{\mathsfit}{\encodingdefault}{\sfdefault}{m}{sl}
\SetMathAlphabet{\mathsfit}{bold}{\encodingdefault}{\sfdefault}{bx}{n}

\usepackage{amsmath,amsfonts,bm}

\def\eqref#1{equation~\ref{#1}}
\def\1{\bm{1}}
\DeclareMathAlphabet{\mathsfit}{\encodingdefault}{\sfdefault}{m}{sl}
\SetMathAlphabet{\mathsfit}{bold}{\encodingdefault}{\sfdefault}{bx}{n}

%% file: requirements/table.tex
\usepackage{multirow}
\usepackage{diagbox}
\usepackage{makecell}
\usepackage{tabularx}
\usepackage{graphicx}

\usepackage{array}
\usepackage{rotating}

\definecolor{aliceblue}{rgb}{0.94, 0.97, 1.0}
\definecolor{citecolor}{HTML}{0071BC}
\definecolor{linkcolor}{HTML}{ED1C24}
\definecolor{darkgreen}{HTML}{539165}

\makeatletter
\newcommand{\thickhline}{%
 \noalign {\ifnum 0=`}\fi \hrule height 1pt
 \futurelet \reserved@a \@xhline
}
\makeatother

\newcommand{\tablesize}{
  \fontsize{8.6pt}{10pt}\selectfont
}

\newcommand{\tablelargesize}{
  \fontsize{9.5pt}{10pt}\selectfont
}

%% file: sections/abstract.tex
\abstract{
In this paper, we propose VideoLLaMA 3, a more advanced multimodal foundation model for image and video understanding.
The core design philosophy of VideoLLaMA3 is vision-centric.
The meaning of ``vision-centric'' is two-fold: the vision-centric training paradigm and vision-centric framework design.
The key insight of our vision-centric training paradigm is that 
\textcolor{cvblue}{\textbf{high-quality image-text data is crucial}} for both image and video understanding. Instead of preparing massive video-text datasets, we focus on constructing large-scale, high-quality image-text datasets. VideoLLaMA3 has four training stages:
1) \textbf{Vision Encoder Adaptation}, which enables the vision encoder to accept images of variable resolutions as input; 2) \textbf{Vision-Language Alignment}, which jointly tunes the vision encoder, projector, and LLM with large-scale image-text data covering multiple types (including scene images, documents, and charts) as well as text-only data. 3) \textbf{Multi-task Fine-tuning}, which incorporates image-text SFT data for downstream tasks and video-text data to establish a foundation for video understanding. 4) \textbf{Video-centric Fine-tuning}, which further improves the model's capability in video understanding.
As for the framework design, to better capture fine-grained details in images, the pretrained vision encoder is adapted to encode images of varying sizes into vision tokens with corresponding numbers, rather than a fixed number of tokens. For video inputs, we reduce the number of vision tokens according to their similarity so that the representation of videos will be more precise and compact.
Benefiting from vision-centric designs, VideoLLaMA3 achieves compelling performances in both image and video understanding benchmarks. Project repository: \url{https://github.com/DAMO-NLP-SG/VideoLLaMA3}.
}

%% file: sections/introduction.tex
\section{Introduction}

% In recent years, the field of artificial intelligence (AI) has achieved significant advancements~\citep{openai2023gpt4tr,team2023gemini,anthropic2024claude3}, profoundly transforming industries and societal functions across the board. 
% Models capable of image recognition~\citep{bai2023qwen,dong2024internlm,chen2023internvl,liu2024visual,chen2023sharegpt4v,lin2023sphinx,young2024yi} and photorealistic image generation~\citep{esser2024scaling,saharia2022photorealistic,ramesh2021zero} have approached near-human capabilities, catalyzing major breakthroughs in sectors such as medical imaging~\citep{li2024llava,tu2024towards} and autonomous driving~\citep{xu2023drivegpt4,jin2023adapt}. 
% Despite these successes, the domain of video understanding and generation~\citep{kondratyuk2023videopoet,menapace2024snap,kondratyuk2023videopoet,videoworldsimulators2024} remains relatively nascent. Unlike static images, videos incorporate temporal dynamics and synchronous audio streams, significantly enriching the information content.
% This integration of continuous audio-visual data complicates extracting and interpreting meaningful patterns, amplifying data complexity and introducing unique computational challenges. 

% background of VLMs

Recent years have witnessed the rapid growth of Large Language Models (LLMs)~\cite{openai2024gpt4o,TheC3,team2024gemini,llama3,yang2024qwen2,liu2024deepseek}, which significantly enhance natural language processing and understanding. The growth of LLMs enables intelligence at the language level. However, to progress further, we need intelligence that extends beyond language, as the world itself is inherently multimodal. Specifically, the model should be capable of perceiving both static scenes and dynamic environments, which necessitates the ability to understand images and videos. Building upon the success of LLMs, Multimodal LLMs (MLLMs)~\cite{wang2024qwen2,aria,wu2024deepseek,chen2024expanding} have been proposed for multimodal understanding.

\begin{figure}[t]
    \centering
    \includegraphics[width=\linewidth]{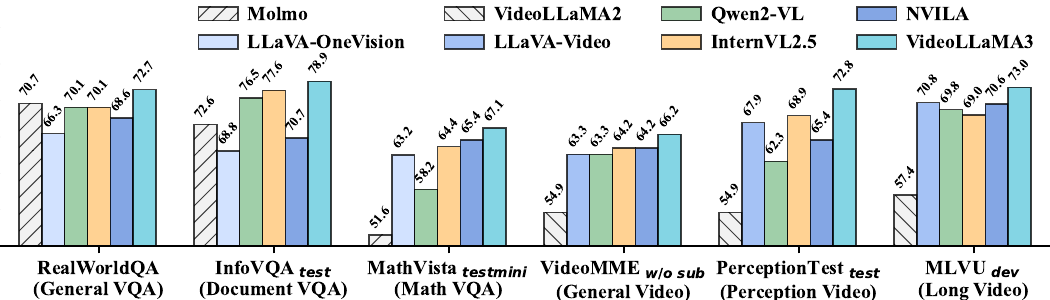}
    \caption{\textbf{Performance Comparison} of VideoLLaMA3 with the previous advanced image/video MLLM on various representative benchmarks. Specifically, VideoLLaMA3 not only demonstrates strong video understanding capabilities~(VideoMME, PerceptionTest, MLVU) but also maintains excellent document comprehension abilities~(DocVQA) and multimodal mathematical reasoning skills~(MathVista). 
    Note that LLaVA-OneVision is only used to evaluate image benchmarks, while LLaVA-Video is only used to evaluate video benchmarks.
    }
    \label{fig:teaser}
\end{figure}

% brief review of image VLMs and video VLMs
% \textcolor{cvblue}{\textbf{@Sicong [brief review of image VLMs] + [brief review of video VLMs]}}
% Recent advancements in image and video vision-language models (VLMs) 
% Existing MLLMs have led to significant progress in multimodal understanding.

Existing MLLMs \cite{li2023videochat,lin2023video,ataallah2024minigpt4,zhang2024longva,zohar2024apollo,song2024moviechat+,fei2024video,liu2024kangaroo,liu2024oryx,wang2024internvideo2,jung2024pegasus,Maaz2024VideoGPT+,cheng2024videollama,sun2024video,zhang2024video,Maaz2023VideoChatGPT,li2023mvbench,chen2024sharegpt4video,li2024llavaonevision,mplug3,chen2023internvl,longllava,internlmxcomposer25} have made significant progress in multimodal understanding.
Image-centric MLLMs~\cite{wang2024qwen2,li2024llavaonevision,internvl2.5,chen2023internvl,molmo2024,liu2024nvila,smolvlm2023}, leveraging high-quality image-text datasets~\cite{li2024llava,li2024llavaonevision,tong2024cambrian,mathew2021docvqadatasetvqadocument,laurençon2024building,masry2022chartqa,huang-etal-2024-chart} that are easier to collect and curate, have demonstrated strong performance in image understanding, such as visual question answering, OCR, and document understanding. 
% These models also enable reasoning about image content using natural language\lx{Note: this sentence can be deleted}.
Beyond static content such as images, video-centric MLLMs~\cite{zhang2024video,chen2024sharegpt4video,cheng2024videollama,damonlpsg2023videollama} must tackle the added complexity of modeling the temporal dimension of videos, requiring models to handle dynamic content and capture dependencies across frames. This temporal complexity, combined with the need for large-scale video-text datasets that are often of lower quality and harder to annotate, makes video MLLMs more challenging. 
% Despite these difficulties, video VLMs [citations] have demonstrated progress in tasks like action recognition and event understanding.
These challenges underscore the advantages of using image understanding as a foundation for video understanding. By extending the robust visual capabilities of image MLLMs, video models can focus on and better address the unique challenges of temporal and dynamic content modeling.
% By focusing on enhancing image understanding, one can provide a more solid and efficient foundation for improving video models, ensuring both higher data quality and better performance across multimodal tasks.

% our motivation, two key designs
Inheriting from VideoLLaMA~\cite{DBLP:conf/emnlp/ZhangLB23} and VideoLLaMA2~\cite{damonlpsg2024videollama2}, VideoLLaMA3, a more advanced multimodal foundation model, is proposed for image and video understanding.
We design VideoLLaMA3 in a vision-centric way. Specifically, we propose a vision-centric training paradigm and vision-centric framework designs.
For the training paradigm, considering the intrinsic relationship between image and video modalities - where videos are essentially sequences of temporally correlated images, we prioritize the improvement of image understanding, which in turn enhances the performance of video understanding.
Moreover, compared to video-text data, image-text data is easier to collect and ensures higher data quality.
For vision-centric framework designs, we propose adapting the vision encoder to handle images of any resolution during the image understanding enhancement stage and tuning the encoder to efficiently embed video inputs.
% Motivated by these, we design the VideoLLaMA3 in a vision-centric way. We propose a vision-centric training paradigm and vision-centric framework designs.
% We shift our focus from preparing massive video-text data to constructing high-quality image-text data. 

\begin{figure}[t] 
	\centering
	\includegraphics[width=0.90\linewidth]{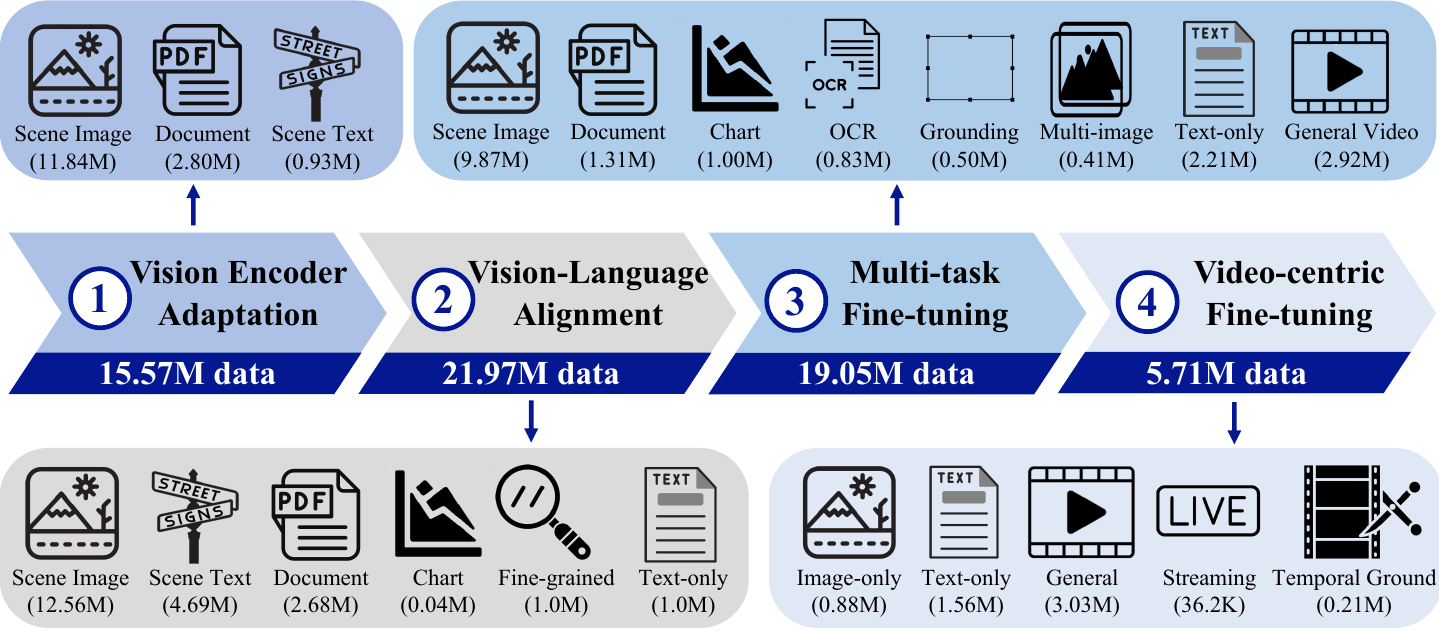} 
	\caption{\textbf{Training paradigm of VideoLLaMA3.} The training of VideoLLaMA3 has four stages: (1) Vision Encoder Adaptation, (2) Vision-Language Alignment, (3) Multi-task Fine-tuning, and (4) Video-centric Fine-tuning.}
	\label{fig:training_pipeline}
\end{figure}

% training paradigm
Our vision-centric training paradigm consists of four stages~(Figure~\ref{fig:training_pipeline}):
\textbf{1) Vision Encoder Adaptation}: This stage aligns the vision encoder's feature space with LLMs. Inputs to the vision encoder are adapted from fixed to dynamic resolutions. Scene images with short captions are used to enhance the encoder's performance, while document and scene text images are used to enable the encoder to capture fine-grained visual details.
\textbf{2) Vision-Language Alignment}: This stage establishes the foundation for multimodal understanding using detailed image-text data. Scene images are annotated with detailed captions, and document and chart data include extensive explanations. To enhance spatial reasoning, fine-grained image-text data with bounding boxes are utilized. A small amount of text-only data is included to retain the model’s language capabilities. All parameters are unfrozen during this stage.
\textbf{3) Multi-Task Fine-tuning}: In this stage, the model is fine-tuned for downstream tasks, such as interactive question answering. Image-text data with questions and answers are employed, along with general video caption data to prepare the model for video perception. The use of general video caption data also surprisingly improves the performance of image understanding.
\textbf{4) Video-centric Fine-tuning}: This final stage enhances the model's performance in video understanding and video question answering. Training data includes general videos, streaming videos, videos annotated with temporal grounding information, image-only and text-only data.

% model design
On the model side, we enhance the vision encoder with two vision-centric designs: 1) we adapt the vision encoder to take images with dynamic resolutions as inputs, and 2) we lift the vision encoder to receive videos and compress the video tokens into more compacted representations.
% As for vision-centric framework designs, we propose two dedicated components:
% We propose two dedicated designs from vision-centric perspectives: 
% 1) vision encoder for visual inputs with dynamic resolutions and 2) vision 
% compressing vision tokens for videos into more compacted representations.
In previous methods~\cite{li2024llavaonevision,liu2023visualit,liu2023improved,internvl2.5,chen2023internvl}, vision tokens are either with fixed numbers or with numbers among several fixed choices, which is an inflexible and unnatural way to represent images. To alleviate this limitation, we adapt the pretrained vision encoder to receive images with variable shapes. This is achieved by replacing the fixed positional embeddings with the Rotary Position Embedding (RoPE). We finetune the vision encoder in the vision encoder adaptation stage so that it can accommodate dynamic inputs. In this way, enabling it to process high-resolution images and images with unusual aspect ratios with minimal information loss.
As for video inputs, we consider the redundant information in videos and propose to reduce the number of vision tokens to represent a video. The advantages of vision token compression are two-fold. One is to make the visual embeddings of videos more compact and precise so that the model can focus more on the dynamic parts of videos. The other is to save computation demands during training and inference for video understanding.

% overview of results
Thanks to the vision-centric training paradigm and framework designs, our proposed VideoLLaMA3 achieves state-of-the-art performance on both image and video understanding benchmarks~(Figure~\ref{fig:teaser}). Notably, in image understanding, the performance in chart understanding and vision-related math problems surpasses state-of-the-art models by a large margin. While in video understanding, our model achieves state-of-the-art performance in most benchmarks including general video understanding, long video understanding, temporal reasoning and grounding.

% summary of contribution
To summarize, the key contributions of VideoLLaMA3 include:
\begin{itemize}
    \item We propose VideoLLaMA3, a more advanced multimodal foundation model, for both image and video understanding. The model achieves state-of-the-art performance on most image and video understanding benchmarks. Notably, VideoLLaMA3 has significant improvements compared to previous versions of VideoLLaMA.
    \item We propose the vision-centric training paradigm. Specifically, we propose to improve video understanding capabilities through large-scale image understanding pretraining.
    \item We propose two vision-centric framework designs to adapt vision encoders to represent images and videos better.
\end{itemize}

%% file: sections/method.tex
\section{Methodology}

\begin{figure*}[tbp] 
\centering
\includegraphics[width=0.84\linewidth]{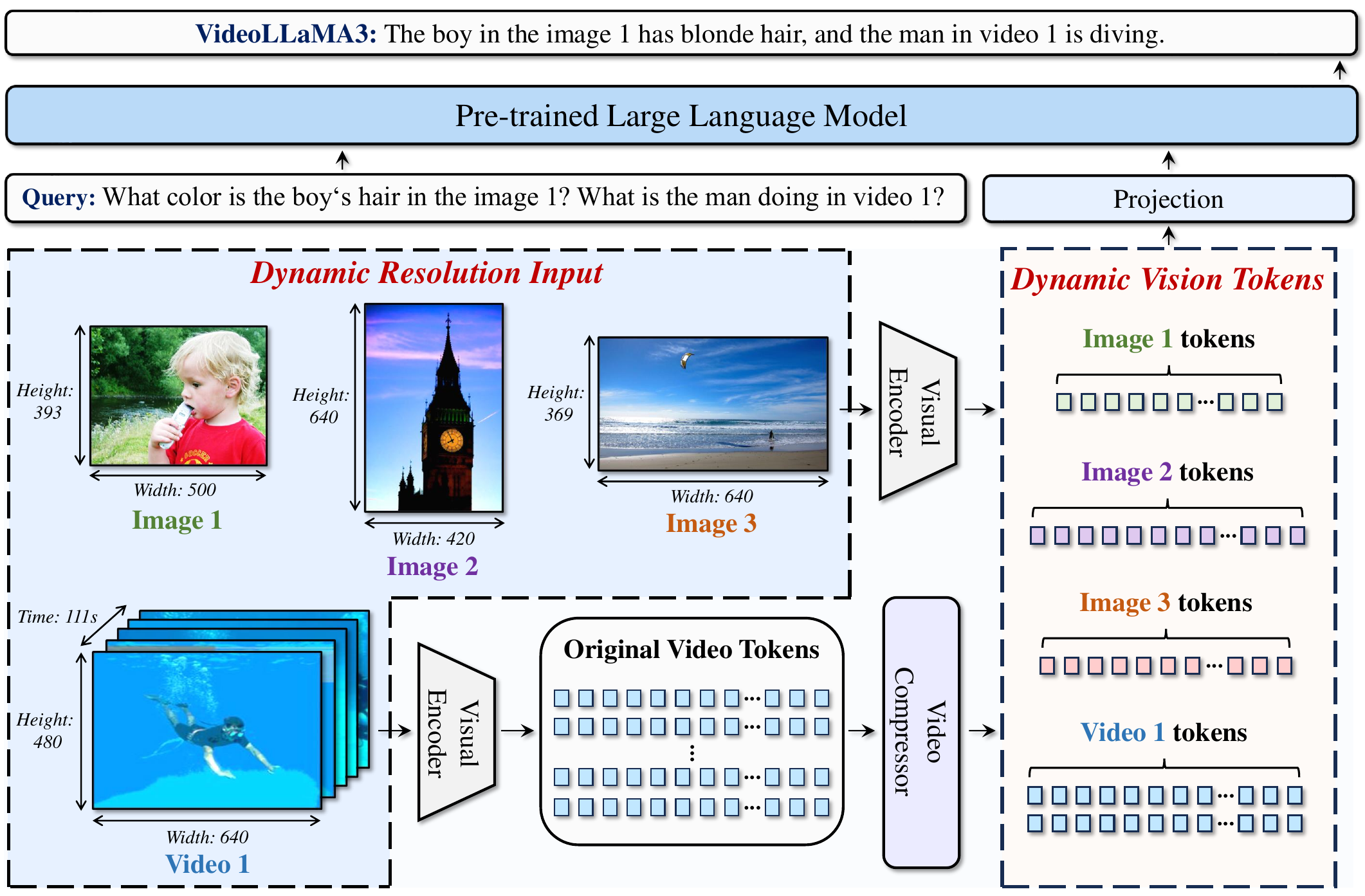} 
\caption{\textbf{The overall pipeline of our VideoLLaMA3.} There are two key technical points: \listnumber{\one} \textbf{Any-resolution Vision Tokenization~(AVT)}: AVT converts images or videos of any resolution into a set of 1-D token sequences, enabling compatibility with varying amounts of input images and videos of different resolutions, thereby supporting more flexible vision input; \listnumber{\two} \textbf{Differential Frame Pruner~(DiffFP)}: Serving as a video compressor, DiffFP eliminates video content with minimal differences between adjacent frames. This approach enhances video processing efficiency, particularly for long-form videos.}
\label{fig:overview_pipeline}
\end{figure*}

As shown in Figure~\ref{fig:overview_pipeline}, on the model side, VideoLLaMA3 consists of two key technical points: \listnumber{\one} \textbf{Any-resolution Vision Tokenization~(AVT)} and \listnumber{\two} \textbf{Differential Frame Pruner~(DiffFP)}.
When it comes to data, since we propose to improve video understanding capabilities based on image understanding, we also develop a pipeline for constructing high-quality re-captioned image dataset.

\subsection{Any-resolution Vision Tokenization}
%Unlike VideoLLaMA~\citep{DBLP:conf/emnlp/ZhangLB23} and VideoLLaMA2~\citep{damonlpsg2024videollama2}, 
In MLLMs, visual inputs are extracted into vision tokens for multimodal understanding. The common practice~\cite{liu2023visualit,liu2023improved} is to extract visual inputs with a pre-trained ViT-based vision encoder. The pre-trained vision encoder only receives images with fixed resolutions, which introduces information loss. To alleviate information loss, AnyRes techniques~\cite{li2024llavaonevision,internvl2.5,chen2023internvl} are proposed to split images into patches with fixed resolutions. Although AnyRes techniques increase the number of vision tokens, it is still inflexible and neglects the position relationship within an image when extracting vision tokens.
% In previous MLLMs~\cite{li2024llavaonevision,,internvl2.5,chen2023internvl}, 
In VideoLLaMA3, we adopt the idea of Any-resolution Vision Tokenization (AVT)~\cite{dehghani2023patch,wang2024qwen2} to dynamically process images and videos of any resolution.
% VideoLLaMA3 can process images and videos of any resolution.
Concretely, we adapt the pre-trained vision encoder (ViT-based architectures) to handle variable resolutions by employing a strategy to replace the absolute position embeddings in ViT with 2D-RoPE~\cite{su2024roformer}.
With AVT, images and videos of different resolutions are better represented with more details included in vision tokens.
To make the vision encoder compatible with AVT, we fine-tune the vision encoder and the projector in the stage of Vision Encoder Adaptation (i.e., stage \#1 in Figure~\ref{fig:training_pipeline}) using scene images, document data, and scene images with texts.
% The vision tokens are thien 
% By using Any-resolution Vision Tokenization (AVT), images and videos of different resolutions are merged into a single sequence.
% Inspired by Qwen2-VL~\citep{qwen2-vl}, since the original ViT used absolute position embeddings that cannot handle variable resolutions, we replaced them with 2D-RoPE. 
% Subsequently, we continued to fine-tune the ViT to capture dynamic positional information. 
% Similarly to Qwen2-VL~\citep{qwen2-vl}, we used a simple MLP layer after ViT to reduce vision tokens by merging adjacent 2 × 2 tokens into one. By using Any-resolution Vision Tokenization (AVT), images and videos of different resolutions are merged into a single sequence.

\begin{figure*}[tbp] 
\centering
\includegraphics[width=1.\linewidth]{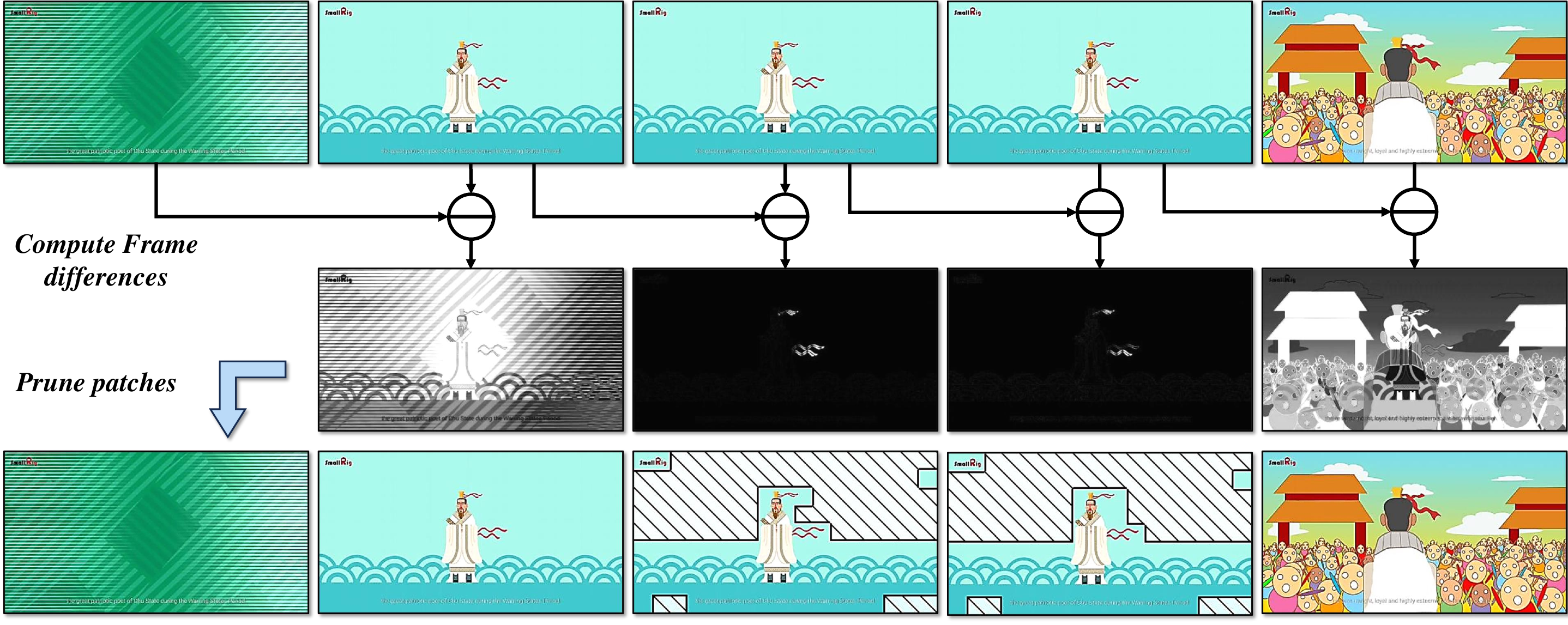} 
\caption{\textbf{The calculation flow of our DiffFP.} We prune video tokens based on patch similarities in pixel space, removing patches with smaller distances to the previous frame.}
\label{fig:video_compressor}
\end{figure*}

\subsection{Differential Frame Pruner}

For videos, inputs usually have much more tokens than image inputs after tokenization. To reduce the computation demand for videos, we apply a per-frame 2$\times$2 spatial downsampling by bilinear interpolation to limit the context length within a certain range.
Besides, considering that videos consist of frames with overlapping content, representing videos by stacking vision tokens from each frame leads to lengthy and redundant tokens.
To further reduce the number of tokens of videos, we propose the Differential Frame Pruner (DiffFP) to prune the video tokens. Inspired by RLT~\citep{choudhury2024don}, we compare the 1-norm distance between temporally consecutive patches within the pixel space. We consider temporally consecutive patches with smaller distances to be redundant, and the later patches can be pruned. Specifically, as shown in Figure~\ref{fig:video_compressor}, we first calculate the 1-norm distance between consecutive frames in the pixel space and then remove patches whose distances fall below a pre-defined threshold. Following RLT~\citep{choudhury2024don}, we set the default threshold to 0.1.

\subsection{Construction of High-Quality Image Re-Caption Dataset}
To train our VideoLLaMA3, we constructed a high-quality image re-caption dataset, VL3-Syn7M. All images in this dataset are sourced from COYO-700M~\citep{kakaobrain2022coyo-700m} and processed using our proposed cleaning pipeline as below:
% . The primary steps are as follows:

\noindent\textbf{1) Aspect Ratio Filtering.} We begin by filtering images based on their aspect ratios, removing those with extreme values. This step ensures that the dataset contains images with typical aspect ratios, preventing potential biases during feature extraction. For instance, images that are excessively long or wide may distort the model's interpretation due to their unusual shapes.

\noindent\textbf{2) Aesthetic Score Filtering.} An aesthetic scoring model is applied to evaluate the visual quality of the images. Based on these scores, images with low aesthetic ratings are discarded. This step eliminates visually poor or poorly composed images, reducing noise and improving the quality of the descriptions generated by the model.

\noindent\textbf{3) Text-Image Similarity Calculation with Coarse Captioning.} The BLIP2 model is used to generate initial captions for images, followed by calculating the text-image similarity using the CLIP model. Images with low similarity are excluded, as they are likely to contain content that is challenging to describe concisely. This process ensures that the remaining images are both descriptive and interpretable.

\noindent\textbf{4) Visual Feature Clustering.} Visual features are extracted using the CLIP vision model, and a k-Nearest-Neighbors (KNN) algorithm is applied for clustering. This method identifies cluster centers in the visual feature space. From each cluster, we select a fixed number of images. This approach ensures diversity within the dataset while maintaining a balanced distribution of semantic categories, improving the model's ability to generalize across various visual content.

\noindent\textbf{5) Image Re-caption.} After filtering and clustering the images, we proceed with detailed re-captioning. Brief captions are generated using InternVL2-8B~\cite{chen2023internvl,Chen2024HowFA}, while the detailed captions are produced with InternVL2-26B~\cite{chen2023internvl,Chen2024HowFA}. These two types of captions (VL3-Syn7M-short and VL3-Syn7-detailed) are employed at different stages of training to address varying needs.

Through the aforementioned cleaning and re-caption process, we created the VL3-Syn7M dataset, which consists of 7 million image-caption pairs. This high-quality dataset will be a crucial component in training our model, providing a rich and diverse set of images and annotations that support strong performance across a wide range of visual tasks.

%% file: sections/data_mixture.tex
\section{Training}
% \lx{Note: I think we should briefly introduce the building blocks of our VideoLLaMA3 here}

As illustrated in Figure~\ref{fig:overview_pipeline}, VideoLLaMA3 consists of four key components: a vision encoder, a video compressor, a projector, and a large language model (LLM). The vision encoder extracts visual tokens and is initialized with the pre-trained SigLIP~\citep{Zhai2023SigmoidLF}. To reduce the number of vision tokens representing videos, a video compressor is employed. The projector bridges the features between the vision encoder and the LLM. For the LLM, we utilize Qwen2.5 models~\cite{yang2024qwen2}.

Inspired by previous explorations in MLLMs~\cite{wang2024qwen2, li2024llavaonevision, zhang2024video}, we develop video understanding capabilities based on strong image understanding foundations.
To enable the model with strong image and video understanding capabilities simultaneously,
the training of VideoLLaMA3 has four stages:
% 
% To train a comprehensive MLLM for image and video understanding based on a pre-trained large language model, we divide the training process into four stages. As illustrated in Figure~\ref{fig:training_pipeline}, these stages are: 
1) Vision Encoder Adaptation, 2) Vision-Language Alignment, 3) Multi-task Fine-tuning, and 4) Video-centric Fine-tuning. 
While the first three stages primarily focus on improving image understanding, the final stage is dedicated to video understanding. The details of the training stages are as follows:
% Following the idea outlined in \citep{li2024llavaonevision}, \ie ``quality over quantity'', we conducted rigorous cleaning procedures to guarantee data quality.
% , which enabled our model to achieve exceptional performance despite using a relatively small dataset.

\noindent\textbf{1) Vision Encoder Adaptation.} In this stage, we fine-tune the vision encoder, which is initialized with the pre-trained SigLIP~\citep{Zhai2023SigmoidLF}, on a large-scale image dataset. During this stage, the vision encoder is made trainable, while the language decoder remains frozen. This fine-tuning transforms the encoder into a dynamic-resolution processor, enhancing its ability to process images of varying resolutions. Meanwhile, the projector is trained to better align the features of the vision encoder with those of the LLM.
% In the Vision Encoder Adaptation stage, we fine-tune the vision encoder, which is initialized with the pre-trained SigLip~\citep{Zhai2023SigmoidLF}, on a large-scale image dataset to transform it into a dynamic-resolution encoder\lx{Note: rewrite this sentence, the details are not clearly presented}\lx{Note: you should point out that the vision encoder is trainable while the language decoder is kept frozen in this stage}. This transformation enhanced the model’s ability to process images at varying resolutions, improving its visual representation capabilities and enabling it to handle a broader range of visual inputs with greater accuracy and efficiency.

\noindent\textbf{2)Vision-Language Alignment.} This stage primarily focuses on introducing multimodal knowledge into the model. During this phase, all parameters are made trainable, enabling both the LLM and the vision encoder to be fine-tuned for integrating multimodal knowledge.

% In the subsequent Vision-Language Alignment Stage, we incorporated multimodal knowledge into the model using a carefully curated, high-quality dataset that included a variety of images labeled with detailed textual descriptions. This process allowed the model to acquire world knowledge essential for interpreting real-world scenes, including recognizing common objects, understanding spatial relationships, and making inferences about the context of visual inputs. Additionally, the model developed robust optical character recognition (OCR) capabilities and fine-grained understanding skills, which enabled it to extract textual information from images and understand complex visual details. 

\noindent\textbf{3) Multi-task Fine-tuning.} In this stage, we perform instruction fine-tuning using a diverse set of multimodal question-answering data, which includes both image and video-based questions. This step is crucial for improving the model’s ability to follow natural language instructions and enhancing its multimodal understanding. Moreover, this stage lays the foundation for the model’s video understanding capabilities, enabling it to process and analyze temporal information. Also, in this stage, we introduce the video compressor to reduce the number of video tokens. 

% During the Multi-task Fine-tuning Stage, we performed instruction fine-tuning using a diverse set of multimodal question-answering data, which included both image and video-based questions. This step was crucial for improving the model’s ability to follow natural language instructions and enhancing its multimodal understanding, allowing it to better integrate information from both visual and textual sources. Moreover, this stage activated the model’s video understanding capabilities, enabling it to process and analyze temporal information, a key component for understanding dynamic scenes in videos. By the end of this stage, the training of our stable version of the image understanding model was completed, with the model having acquired both strong image understanding and sufficient video understanding capabilities.

\noindent\textbf{4) Video-centric Fine-tuning.} In this stage, we focus on enhancing the model’s video understanding capabilities. All parameters are unfreezed during this stage.
The data used in this stage includes video-text data, image-only data and text-only data.
% The 
% Finally, in the Video-centric Fine-tuning Stage, we focused specifically on optimizing the model’s video understanding capabilities. 
% We created multimodal video instruction fine-tuning data, which covered a wide variety of real-world scenarios and scene categories, from everyday activities to complex, multi-object interactions. Meanwhile, we reduced visual feature size using bilinear interpolation at this stage, which helped to improve processing speed. In addition, we introduced our proposed inter-frame compression technique, which significantly enhanced inference efficiency by reducing the amount of redundant information between continuous frames. This resulted in a highly efficient and powerful multi-scene video understanding model, capable of processing more frames for deeper analysis across a wide range of contexts and delivering accurate insights.

\subsection{Data Format}
\label{sec:data_format}

\begin{figure}[t]
\centering
\includegraphics[width=\linewidth]{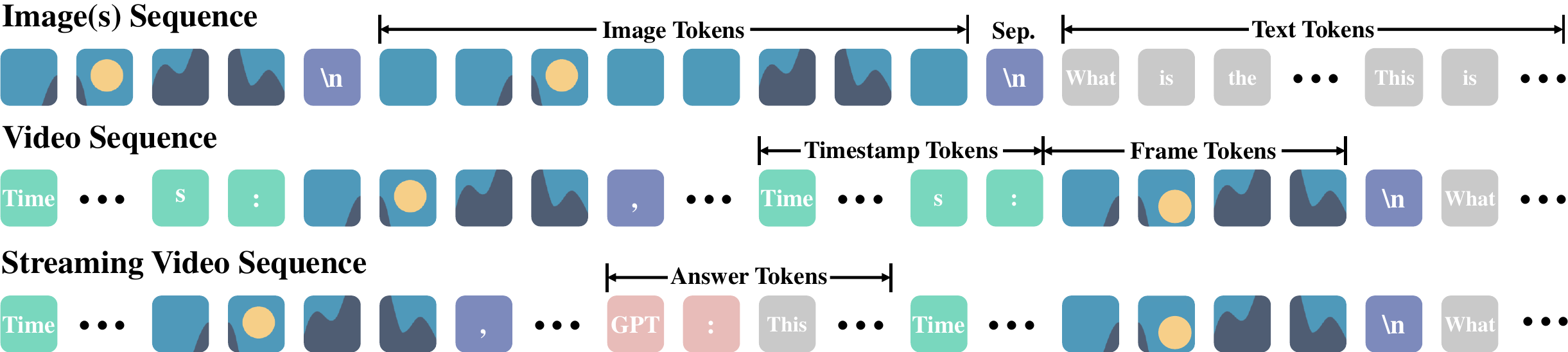}
\caption{\textbf{Data formats for different data types.} \listnumber{\one} For image sequence, we use "\textbackslash n" to separate image tokens from different image; \listnumber{\two} For video sequence, we use "Time: xxs" to indicate timestamps of each frame, "," to separate different frames, and "\textbackslash n" to separate tokens from different videos; \listnumber{\three} For streaming video sequence, videos and texts are organized in an interleaved format.
}
\label{fig:data_format}
\end{figure}

The data format for images, videos and streaming videos are shown in Figure~\ref{fig:data_format}.

\noindent\textbf{Image Sequence.} Images are represented as a sequence of tokens, referred to as Image Tokens. The ``\textbackslash n'' character is used to separate tokens belonging to different images. Besides, text tokens follow image tokens, separated by ``\textbackslash n'', enabling a mixed representation of image and textual data.

\noindent\textbf{Video Sequence.} Frames in a video sequence are represented as Frame Tokens. Before tokens for each frame, a Timestamp Token in the format "Time: xxs" is inserted to denote the time corresponding to that frame. Frames within a video sequence are separated by commas ",".
After the video tokens, ``\textbackslash n'' is inserted to separate the video data from any subsequent text tokens, ensuring a clear distinction between the two modalities.

\noindent\textbf{Streaming Video Sequence.} For streaming video data, video and text tokens are interleaved in the sequence. Timestamps (\ie ``Time: xxs'') are inserted before the frame tokens, similar to video sequences. To mimic the interactive scenarios of streaming videos, Answer tokens (\ie ``GPT: xxx'') may appear within the sequence to denote contextualized outputs or interactions. The interleaved format ensures a seamless integration of video and textual data streams.

\subsection{Data Mixture}
Following the principle outlined in LLaVA-OneVision \citep{li2024llavaonevision}, \ie ``quality over quantity'', we conduct rigorous cleaning procedures to guarantee data quality. In this section, we provide a detailed description of the data mixture for each stage, as well as the synthesis and cleaning methods applied to different data subsets. 
% The correct application of these methods is essential for ensuring the quality and diversity of the training set, which directly influences the model's final performance.

% \subsubsection{Adaptation to Dynamic Resolution}
\subsubsection{Vision Encoder Adaptation}

\begin{table}[ht]
    \tablesize
    \centering
    \renewcommand{\arraystretch}{1.4}
    \caption{\textbf{Data mixture in vision encoder adaptation stage.}}
    \begin{tabularx}{\textwidth}{ p{2.2cm} X >{\centering\arraybackslash}p{1.2cm} }
        \thickhline
        \textbf{Task} & \textbf{Dataset} & \textbf{Amount} \\
        \thickhline

        Scene Image &
        VL3-Syn7M-short, LLaVA-Pretrain-558k~\citep{liu2023improvedllava}, Objects365-Recap~\citep{Objects365}, SA-1B-Recap~\citep{kirillov2023segment} &
        11.84M \\

        \rowcolor[HTML]{EBEBEB} 
        Scene Text Image &
        BLIP3-OCR-Recap~\citep{Xue2024xGenMMA} &
        0.93M  \\

        Document &
        pdfa-eng-wds~\citep{pdfa}, idl-wds~\cite{idlwds} &
        2.80M  \\

        \thickhline
    \end{tabularx}
    \vspace{1pt}
    \label{tab:stage1mixture}
\end{table}

The Vision Encoder Adaptation stage is designed to enhance the model’s ability to comprehend a wide range of diverse scenes and improve its feature extraction capacity, with a particular focus on capturing fine-grained information such as objects, regions, and text. As shown in Table~\ref{tab:stage1mixture}, the training data in this stage combines scene images and document recognition images, along with a small portion of scene text images. It should be noted that all data labeled as "Recap" consists of captions generated with InternVL2-8B~\citep{chen2023internvl}.

For scene images, our data sources include VL3-Syn7M-short, LLaVA-Pretrain-558K~\citep{liu2023improvedllava}, Object365~\citep{Objects365}, and SA-1B~\citep{kirillov2023segment}. Notably, the Object365 and SA-1B datasets are included to enhance data diversity, as images in this dataset are mainly complex scenes. 
% , combined with the original image annotations.

The scene text images are sourced from BLIP3-OCR~\citep{Xue2024xGenMMA}. Both the brief recaption and the text content within the images are used as captions, and the text content caption following a left-to-right, top-to-bottom pattern across the image.

The document images used in this stage are a subset of pdfa-eng-wds~\citep{pdfa} and idl-wds~\citep{idlwds}. 
% Images were selected based on similarity to avoid redundancy in document content. 
A total of 2.8 million images were chosen from these two datasets, with the text content of the documents serving as image captions, following the reading order.

\subsubsection{Vision-Language Alignment}

\begin{table}[h]
    \tablesize
    \centering
    \renewcommand{\arraystretch}{1.4}
    \caption{\textbf{Data mixture in vision-language alignment stage.}}
    \begin{tabularx}{\textwidth}{ p{2.2cm} X >{\centering\arraybackslash}p{1.2cm} }
        \thickhline
        \textbf{Task} & \textbf{Dataset} & \textbf{Amount} \\
        \thickhline

        Scene Image &
        VL3-Syn7M-detailed, Objects365-Recap~\citep{Objects365}, SA-1B-Recap~\citep{kirillov2023segment}, COCO2017-Recap~\citep{lin2014microsoft}, ShareGPT4o~\citep{Chen2024HowFA}, TextCaps~\citep{sidorov2020textcaps}, ShareGPT4V~\citep{chen2023sharegpt4v}, DenseFusion~\citep{li2024densefusion}, LLaVA-ReCap (LCS-558K)~\citep{li2024llavaonevision}
        & 12.56M \\

        \rowcolor[HTML]{EBEBEB} 
        Scene Text Image &
        Laion-OCR~\citep{schuhmann2022laion}, COCO-Text~\citep{veit2016coco}, TextOCR~\citep{singh2021textocr}, BLIP3-OCR-Recap~\citep{Xue2024xGenMMA}, LSVT~\citep{Sun2019ICDAR2C}, ReCTS~\citep{liu2019icdar} &
        4.69M  \\

        Document &
        SynthDoG-EN~\citep{kim2022ocr}, SynthDoG-ZH~\citep{kim2022ocr}, UReader-TR~\citep{Ye2023UReaderUO}, FUNSD~\citep{funsd}, DUDE~\citep{van2023icdar}, Vary-600k~\citep{wei2023vary}, pdfa-eng-wds~\citep{pdfa}, idl-wds~\citep{idlwds} &
        2.68M  \\

        \rowcolor[HTML]{EBEBEB} 
        Chart &
        Chart-to-Text~\citep{2022Chart} &
        0.04M  \\

        Fine-grained &
        Osprey-724K~\citep{yuan2024osprey}, MDVP-Data~\citep{lin2024draw}, ADE20K-Recap~\citep{zhou2019semantic}, Object365~\citep{Objects365}, Flickr-30K~\citep{young2014image}, GranD~\citep{rasheed2024glamm} &
        1.00M  \\
    
        \rowcolor[HTML]{EBEBEB} 
        Text-only & 
        Evol-Instruct-143K~\citep{chen2024allava}, Infinity-Instruct-code~\citep{InfinityInstruct2024}, Infinity-Instruct-commonsense~\cite{InfinityInstruct2024}, Infinity-Instruct-math~\citep{InfinityInstruct2024} &
        6.25M  \\
    
        \thickhline
    \end{tabularx}
    % \vspace{1pt}
    \label{tab:stage2mixture}
\end{table}

In this stage, we fine-tune the model using high-quality data.
% , following the acquisition of a powerful vision encoder. 
% Specifically, we leverage large-scale, high-quality datasets to construct question-answering datasets for training. 
As shown in Table~\ref{tab:stage2mixture}, we curate five types of data to cover a wide range of everyday scenarios: scene images, scene text images, documents, charts, and fine-grained data, along with a substantial amount of high-quality text-only data.

For scene images, we include COCO-2017~\citep{veit2016coco}, Object365~\citep{Objects365}, SA-1B~\citep{kirillov2023segment}, ShareGPT4o~\citep{Chen2024HowFA}, ShareGPT4V~\citep{chen2023sharegpt4v}, DenseFusion~\citep{li2024densefusion}, and LLaVA-Recap (LCS-558K)~\citep{li2024llavaonevision}. For Object365, COCO-2017, and SA-1B datasets, we combined the original image annotations with InternVL2-26B~\cite{chen2023internvl} to recaption and generate detailed image captions.

The scene text images include a diverse set of Chinese and English scene text recognition datasets. These datasets, such as BLIP3-OCR~\citep{Xue2024xGenMMA}, COCO-Text~\citep{veit2016coco}, TextOCR~\citep{singh2021textocr}, LSVT~\citep{Sun2019ICDAR2C}, and ReCTS~\citep{liu2019icdar}, provide varied examples of text in real-world environments. 
% Using these data sets, we enable the model to recognize and read text in diverse contexts, such as street signs and shop names, where the text may differ in style, angle, or background. 
Furthermore, we filter images from the LAION dataset~\citep{schuhmann2022laion} to include those with clear and readable text, resulting in a collection of 3 million high-quality images, which we term as Laion-OCR dataset. 
% images from the Laion~\citep{schuhmann2022laion} dataset. 
% After cleaning and filtering, we obtained approximately 3 million high-quality images, which we refer to as Laion-OCR. We use PadlleOCR to extract the texts in these images. 
For the Laion-OCR dataset captions, we include both the text content and the corresponding bounding box annotations of the text locations. The caption format is as follows:

\begin{tcolorbox}[colback=gray!20, colframe=gray!40, coltext=black, boxrule=0.5mm]
\{Caption\}. The texts in this image are \{Text1\}<box>[\{Bounding Box 1\}]</box>, \{Text2\}<box>[\{Bounding Box 2\}]</box>, ...
\end{tcolorbox}

As for document images, we include pdfa-eng-wds~\citep{pdfa}, idl-wds~\citep{idlwds}, UReader-TR~\citep{Ye2023UReaderUO}, Vary-600k~\citep{wei2023vary}, and SynthDoG~\citep{kim2022ocr}.
% 
% When constructing the document image datasets, in addition to the pdfa-eng-wds~\citep{pdfa} and idl-wds~\citep{idlwds} datasets, we also add the high-quality synthetic dataset SynthDoG~\citep{kim2022ocr}. 
SynthDoG dataset is constructed by generating synthetically accurate document images, avoiding human annotation errors and ensuring precise model training. 
% However, synthetic images often differ from real-world images in style and detail. 
% To address this, we incorporate the UReader-TR~\citep{Ye2023UReaderUO} and Vary-600k~\citep{wei2023vary} datasets, which offer more diverse and real image samples with accurate annotations. 
Furthermore, we add the handwritten document dataset FUNSD~\citep{funsd} and the complex document dataset DUDE~\citep{van2023icdar}. FUNSD provides annotated handwritten samples for handwriting recognition, while DUDE includes documents with complex layouts, enhancing the model’s ability to handle a variety of document types. 
% These additions help improve the model's robustness and ensure the training data better reflect real-world document forms.

For chart images, since charts share many similarities with documents in terms of content presentation, 
% particularly in the challenges of fine-grained text recognition and understanding, 
we only include a limited amount of chart data. These data come from the Chart-to-Text~\citep{2022Chart} dataset.

For fine-grained images, we construct two types of data: region caption data and grounded caption data. Region caption data describes the content of specific regions within an image. These data are derived and constructed from the Ospery-724K~\citep{yuan2024osprey}, Object365~\citep{Objects365}, ADE20K~\citep{lin2024draw}, and MDVP-Data~\citep{zhou2019semantic} datasets. These data help the model to understand the details of the image at the region level. Grounded caption data consist of textual descriptions of objects with corresponding bounding box annotations, primarily constructed from the Flickr-30K~\citep{young2014image} and GranD~\citep{rasheed2024glamm} datasets. Both types of data enhance the model's understanding of images, supporting more accurate object localization and recognition in complex scenes.

\subsubsection{Multi-task Fine-tuning}

\begin{table}[h]
    \tablesize
    \centering
    \renewcommand{\arraystretch}{1.4}
    \caption{\textbf{Data mixture in massive multi-task fine-tuning stage.}}
    \begin{tabularx}{\textwidth}{ p{2.2cm} X >{\centering\arraybackslash}p{1.2cm} }
        \thickhline
        \textbf{Task} & \textbf{Dataset} & \textbf{Amount} \\
        \thickhline

        \textit{Image \& Text Data}  \\
        \hline

        General &
        LLaVA-SFT-665K~\citep{li2024llava}, LLaVA-OV-SI~\citep{li2024llavaonevision}, Cambrian-cleaned~\citep{tong2024cambrian}, Pixmo (docs, cap, points, cap-qa, ask-model-anything)~\citep{molmo2024} &
        9.87M \\

        \rowcolor[HTML]{EBEBEB}
        Document &
        DocVQA~\citep{mathew2021docvqadatasetvqadocument}, Docmatix~\citep{laurençon2024building} &
        1.31M \\

        Chart/Figure &
        ChartQA~\citep{masry2022chartqa}, MMC\_Instruction~\citep{liu2023mmc}, DVQA~\citep{kafle2018dvqa}, 
        LRV\_Instruction~\citep{liu2023aligning}, ChartGemma~\citep{masry2024chartgemmavisualinstructiontuningchart}, InfoVQA~\citep{mathew2022infographicvqa}, PlotQA~\citep{methani2020plotqa} &
        1.00M \\

        \rowcolor[HTML]{EBEBEB}
        OCR &
        MultiUI~\citep{liu2024harnessingwebpageuistextrich}, \textcolor{gray}{in-house data} &
        0.83M  \\

        Grounding &
        RefCoco~\citep{kazemzadeh2014referitgame}, VCR~\citep{zellers2019vcr}, \textcolor{gray}{in-house data} &
        0.50M \\

        \rowcolor[HTML]{EBEBEB}
        Multi-Image &
        Demon-Full~\citep{li2024fine}, Contrastive\_Caption~\citep{jiang2024mantisinterleavedmultiimageinstruction} &
        0.41M \\

        Text-only &
        Magpie~\citep{xu2024magpie}, Magpie-Pro~\citep{xu2024magpie}, Synthia~\citep{Synthia-70B-v1.2}, Infinity-Instruct-subjective~\citep{InfinityInstruct2024}, NuminaMath~\citep{li2024numinamath} &
        2.21M \\

        \hline
        \textit{Video \& Text Data} \\
        \hline

        \rowcolor[HTML]{EBEBEB}
        General &
        LLaVA-Video-178K~\citep{zhang2024video}, ShareGPT4o-Video~\citep{chen2024sharegpt4video}, FineVideo~\citep{Farré2024FineVideo}, CinePile~\citep{rawal2024cinepile}, ShareGemini-k400~\citep{sharegemini}, ShareGemini-WebVID~\citep{sharegemini}, VCG-Human~\citep{Maaz2024VideoGPT+}, VCG-Plus~\citep{Maaz2024VideoGPT+}, \textcolor{gray}{VideoLLaMA2 in-house data}, \textcolor{gray}{Temporal Grounding in-house data} &
        2.92M \\

        \thickhline
    \end{tabularx}
    % \vspace{1pt}
    
    \label{tab:stage3mixture}
\end{table}

In this stage, we perform instruction tuning with instruction-following data to refine the model's ability to interpret and follow natural language instructions. 
% To enhance both image understanding and instruction-following capabilities, we construct a comprehensive multi-task data mixture. 
This data mixture is designed to cover a wide range of tasks, enabling the model to learn to perform various actions based on instructions across diverse contexts and modalities. 
Additionally, to activate the model’s video understanding capabilities, we incorporate general video data.

Similar to the vision-language alignment stage, we divide the image data into six distinct groups: general, document, chart/figure, OCR, grounding, and multi-image, as shown in Table~\ref{tab:stage3mixture}. Each category targets at a specific aspect of visual understanding, ensuring the model can effectively handle tasks related to different types of visual information. Alongside these visual data categories, we also include a substantial amount of text-only data to improve the model’s ability to handle diverse instruction-following tasks involving both visual and textual inputs.

The general image data includes high-quality datasets, such as LLaVA-SFT-665K~\citep{liu2023improvedllava} and LLaVA-OV-SI~\citep{li2024llavaonevision}, which serve as foundational resources for enhancing the model’s scene understanding. We also clean and filter the Cambrian-10M~\citep{tong2024cambrian} dataset.
% , which plays an important role in the training process at this stage. 
Furthermore, we incorporate meaningful data from the Pixmo dataset~\citep{molmo2024}, including tasks such as document analysis, caption generation, and counting. These scene images cover a wide range of tasks, including captioning, counting, document understanding, mathematical reasoning, and \textit{etc}.

For constructing the document and chart/figure datasets, we carefully select high-quality data sources and perform quality cleaning to ensure data reliability. It should be noted that the Docmatix dataset is included as it contains multi-page and diverse documents, crucial for significantly enhancing the model’s ability to understand and long complex document structures and content.

For OCR data, we consider two common cases in real-world scenarios: development scenarios and natural scenarios.
% 
% For OCR data, we classify real-world scenarios into two major categories: electronic scenes (e.g., web pages and UI elements) and natural scenes. 
For development scenarios, we use the MultiUI dataset~\citep{liu2024harnessingwebpageuistextrich} to activate the model’s capabilities in understanding and processing text within user interfaces. 
For natural scenarios, we leverage the Laion-OCR dataset to construct additional instruction-tuning data. 
% This newly constructed in-house data further improves the model’s performance in recognizing and interpreting text in natural scene contexts. 
The instruction-tuning data for OCR consists of the following five sub-tasks:
% 
% \begin{tcolorbox}[colback=gray!20, colframe=gray!40, coltext=black, boxrule=0.5mm]
1) Text Existence Detection: Determine whether a specific piece of text exists within the image.
2) Text Localization: Locate a specific piece of text within the image and output its bounding box.
3) Text Recognition within a Bounding Box: Given a bounding box, recognize the text contained within it.
4) Text Comparison Between Images: Given two images, determine in which image the specified text appears.
5) Comprehensive Text Detection and Recognition: Detect and recognize all text present in the image.
% \end{tcolorbox}

For grounding images, we select data from established datasets such as RefCOCO~\citep{kazemzadeh2014referitgame} and VCR~\citep{zellers2019vcr}, which focus on tasks of grounding visual elements in specific textual descriptions.

For multi-image scenes, we leverage the Demon-Full~\citep{li2024fine} and Contrastive-Caption~\citep{jiang2024mantisinterleavedmultiimageinstruction} datasets. The Demon-Full dataset is particularly valuable as it includes various tasks involving multi-image scenes, such as comparing differences between two images, generating captions for the final image in a comic strip, completing missing text in images with occluded portions, determining whether multiple images belong to the same category, and more. These tasks help the model handle complex scenarios involving multiple images, providing a more comprehensive understanding of how visual information can be interpreted across a series of related images. At the same time, such multi-image data further enhances the model's video understanding capabilities.

For the video data used in this stage, we incorporate commonly used high-quality video caption datasets, along with a small amount of question-answering data. In addition, we supplement these with high-quality data from VideoLLaMA2~\citep{damonlpsg2024videollama2} and in-house temporal grounding data. The in-house temporal grounding data specifically focuses on temporal relationships between video frames, enabling the model to grasp the sequence of events and understand the flow of actions across time. These combined data sources contribute to a more robust and nuanced video understanding capability for the model.

\subsubsection{Video-centric Fine-tuning}

\begin{table}[h]
    \tablesize
    \centering
    \caption{\textbf{Data mixture in video-centric fine-tuning stage.}}
    \renewcommand{\arraystretch}{1.4}
    \begin{tabularx}{\textwidth}{ p{2.2cm} X >{\centering\arraybackslash}p{1.2cm} }
        \thickhline
        \textbf{Task} & \textbf{Dataset} & \textbf{Amount} \\
        \thickhline

        General Video &
        LLaVA-Video-178K~\citep{zhang2024video}, ShareGPT4o-Video~\citep{chen2024sharegpt4video}, FineVideo~\citep{Farré2024FineVideo}, CinePile~\citep{rawal2024cinepile}, ShareGemini-k400~\citep{sharegemini}, ShareGemini-WebVID~\citep{sharegemini}, VCG-Human~\citep{Maaz2024VideoGPT+}, VCG-Plus~\citep{Maaz2024VideoGPT+}, VideoRefer~\citep{yuan2024videorefer}, \textcolor{gray}{VideoLLaMA2 in-house data}, \textcolor{gray}{In-house synthetic data} &
        3.03M \\

        \rowcolor[HTML]{EBEBEB}
        Streaming Video &
        ActivityNet~\citep{krishna2017dense}, YouCook2~\citep{zhou2018towards}, Ego4D-narration~\citep{grauman2022ego4d}, Ego4D-livechat~\citep{chen2024videollm} &
        36.2K \\

        Temporal Grounding &
        ActivityNet~\citep{krishna2017dense}, YouCook2~\citep{zhou2018towards}, ViTT~\citep{huang2020multimodal}, QuerYD~\citep{oncescu2021queryd}, HiREST~\citep{zala2023hierarchical}, Charades-STA~\citep{gao2017tall}, Moment-10M~\citep{qian2024momentor}, COIN~\citep{tang2019coin} &
        0.21M \\

        \rowcolor[HTML]{EBEBEB}
        Image-only &
        LLaVA-SFT-665K~\citep{li2024llava}, LLaVA-OV-SI~\citep{li2024llavaonevision} &
        0.88M \\

        Text-only &
        Magpie~\citep{xu2024magpie}, Tulu 3~\citep{lambert2024tulu3} &
        1.56M \\

        \thickhline
    \end{tabularx}
    \label{tab:stage4mixture}
\end{table}

The video-centric fine-tuning stage is designed to tune VideoLLaMA3 to a video specialist and fully unleash its video understanding ability by focusing mainly on large-scale and high-quality video instruction following.
We first collect videos with generally annotated caption, question, and answer from multiple open-source datasets including LLaVA-Video~\citep{zhang2024video}, ShareGPT-4o~\citep{chen2024sharegpt4video}, FineVideo~\citep{Farré2024FineVideo}, CinePile~\citep{rawal2024cinepile}, ShareGemini~\citep{sharegemini}, VideoGPT+~\citep{Maaz2024VideoGPT+} and VideoRefer~\cite{yuan2024videorefer}.
These about 2.7M video-centric conversations eventually form a dataset across various scenes and tasks to serve as examples for teaching the model to understand complex dynamic and static content in videos.

In addition, we further expand the data scale and strengthen the model by synthesizing dense captions and QAs of specific aspects.
Specifically, following the pipeline proposed in \citep{zhang2024video}, we first filter 68K dynamic videos from Panda-70M~\citep{chen2024panda} dataset by optical flow, and then employ Qwen2-VL-72B~\citep{wang2024qwen2} to generate diverse dense captions and QAs for each video from the aspects of temporal understanding, spatial understanding, object description, and time-order understanding.
Finally, 242K question-answer pairs are used for training.

Besides general video-centric conversations, we also introduce the feature of streaming video understanding and temporal grounding to extend the application scenarios of our model.
For streaming video understanding, we acquire data from ActivityNet~\citep{krishna2017dense}, YouCook2~\citep{zhou2018towards}, and Ego4D~\citep{grauman2022ego4d}, and organize video frames and multiple temporal dense captions in an interleaved manner as described in Section~\ref{sec:data_format}, aiming at enhancing the ability to understand fine-grained events in video and to sustain multi-turn conversations in streaming video.
Since these videos are generally long, we cut them into small segments of up to two minutes according to the time interval of dense captions, and remove clips with overly dense and sparse captions.
The synthetic streaming conversation from VideoLLM-Online~\citep{chen2024videollm} is also involved.
For temporal grounding, we collect 205K data from datasets including ActivityNet~\citep{krishna2017dense}, YouCook2~\citep{zhou2018towards}, ViTT~\citep{huang2020multimodal}, QuerYD~\citep{oncescu2021queryd}, HiREST~\citep{zala2023hierarchical}, Charades-STA~\citep{gao2017tall}, Moment-10M~\citep{qian2024momentor}, and COIN~\citep{tang2019coin}, and directly convert the grounding annotation to text format such as ``1.0-2.0 s'' for training.

Finally, we emloy a certain amount of image-only and text-only data from LLaVA~\citep{li2024llava}, LLaVA-OneVision~\citep{li2024llavaonevision}, Magpie~\citep{xu2024magpie}, and Tulu 3~\citep{lambert2024tulu3} for mitigating the impact of catastrophic forgetting on the model’s capabilities.

\subsection{Implementation Details}
% \lx{Note: give the details of the pre-trained models and the parameter settings of each training stage}

In this part, we briefly introduce the implementation details of each training stage. For all stages, we adopt the cosine learning rate scheduler. The warm up ratio of the learning rate is set as $0.03$. The maximum token length is set as $16384$, while the maximum token length for vision tokens is set as $10240$. In the stage of Vision Encoder Adaptation, when training VideoLLaMA3-2B, we initialize the vision encoder with the pre-trained weights of SigLIP~\citep{Zhai2023SigmoidLF} and the LLM with the pre-trained weights of Qwen2.5-2B~\cite{yang2024qwen2}. For VideoLLaMA3-7B, the vision encoder is initialized with the fine-tuned SigLIP weights in VideoLLaMA3-2B and the LLM is initialized with Qwen2.5-7B~\cite{yang2024qwen2}. The projector is implemented as a two-layer MLP with GELU as the activation function.
In this stage, we only train the vision encoder and projector, and their learning rates are set as $1.0 \times 10^{-5}$ and $1.0 \times 10^{-3}$, respectively.
For the remaining stages, the learning rates for the LLM, the projector, and the vision encoder are set as $1.0 \times 10^{-5}$, $1.0 \times 10^{-5}$, $2.0 \times 10^{-6}$, respectively.
The differential frame pruner is applied in the multi-task fine-tuning stage and the video-centric fine-tuning stage where video data is involved. The threshold to discard similar visual tokens is $0.1$.
To limit context length, the visual tokens of videos are spatially downsampled after vision encoder by a factor of $2$ using bilinear interpolation.
The visual tokens of images are only downsampled in the video-centric fine-tuning stage to align with video data.
For loading video data, we first sample frames at $1$ frame per second using FFmpeg.
These frames will be further sampled uniformly if the total number of frames is greater than a certain value, which is set to $180$ to accommodate most videos that last less than $3$ minutes.

%% file: sections/experiment.tex
\section{Experiment}

\begin{table*}[t]
  \tablesize
  \centering
  \renewcommand{\arraystretch}{1.4}
  \caption{\textbf{Evaluation results of 2B models on image benchmarks.} $^*$ denotes the reproduced results. The best results are \textbf{in bold} and the second best ones are \underline{underlined}.}
  \begin{tabularx}{\textwidth}{ p{3.0cm} | >{\centering\arraybackslash}X | >{\centering\arraybackslash}X | >{\centering\arraybackslash}X | >{\centering\arraybackslash}X }
    \thickhline

    \diagbox[width=3.4cm]{\textbf{Benchmark}}{\textbf{Model}} & 
    \makecell{\includegraphics[height=0.38cm]{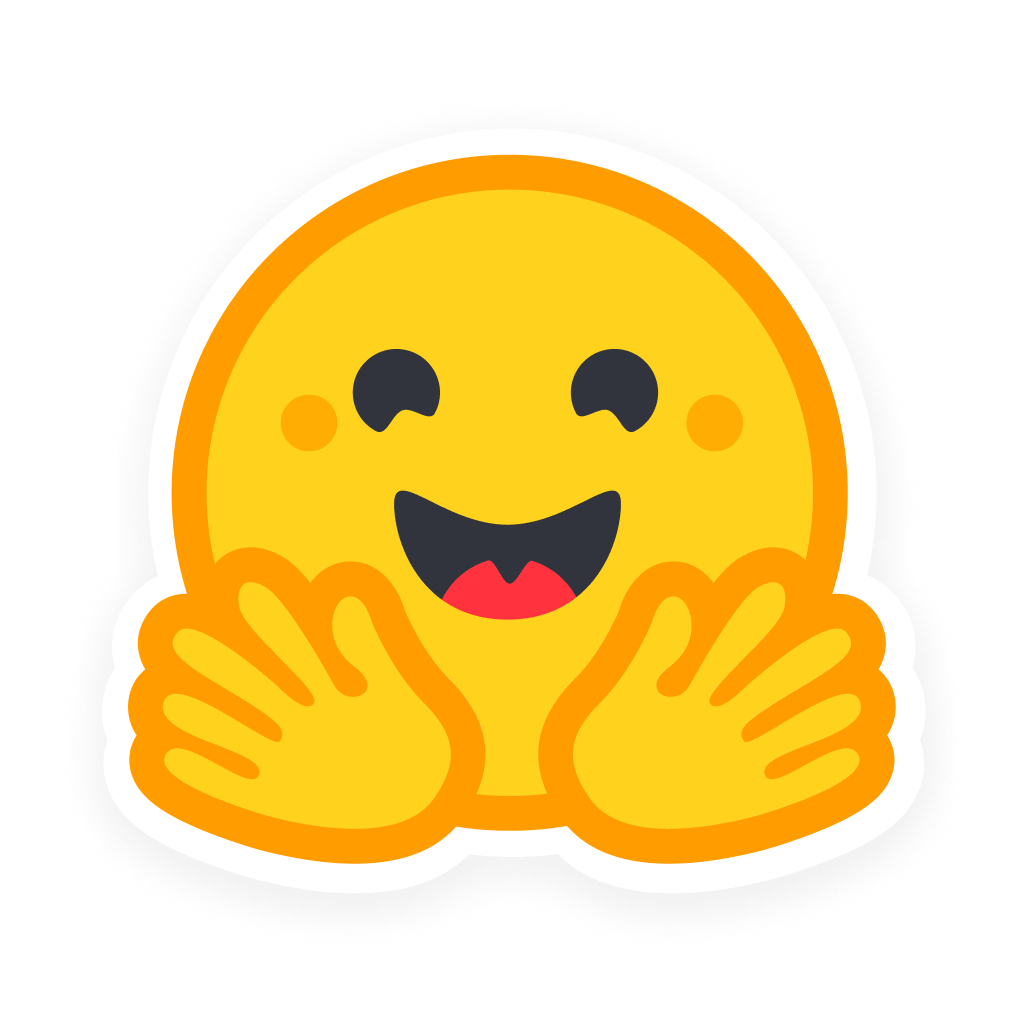} \textbf{SmolVLM} \\ \textbf{2B}} & 
    \makecell{\includegraphics[height=0.3cm]{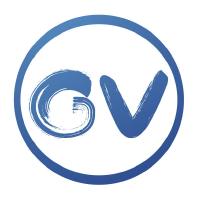} \textbf{InternVL2.5} \\ \textbf{2B}} & 
    \makecell{\includegraphics[height=0.3cm]{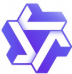} \textbf{Qwen2-VL} \\ \textbf{2B}} & 
    \makecell{\textbf{VideoLLaMA3} \\ \textbf{2B}} \\

    \thickhline

    \rowcolor[gray]{0.9}
    \multicolumn{5}{l}{\emph{Document/Chart/Scene Text Understanding}} \\
    ChartQA & 65.3* & \underline{79.2} & 73.5\phantom{-} & \tablelargesize\textbf{79.8} \\
    DocVQA$_{\mathrm{test}}$ & 81.6 & 88.7 & \underline{90.1}\phantom{-} & \tablelargesize\textbf{91.9} \\
    InfoVQA$_{\mathrm{test}}$ & - & 60.9 & \underline{65.5}\phantom{-} & \tablelargesize\textbf{69.4} \\
    OCRBench & 622* & \textbf{804}\np & \nt767$^*$\np & \tablelargesize\underline{779}\np \\

    % \hline
    % \hline
    \rowcolor[gray]{0.9}
    \multicolumn{5}{l}{\emph{Math}} \\

    MathVista$_{\mathrm{testmini}}$ & 44.6 & \underline{51.3} & 43.0 & \tablelargesize\textbf{59.2} \\
    MathVision$_{\mathrm{test}}$ & 6.5* & \underline{14.7} & 12.4 & \tablelargesize\textbf{15.5} \\

    % \hline
    % \hline
    \rowcolor[gray]{0.9}
    \multicolumn{5}{l}{\emph{Multi Image}} \\

    MMMU-Pro & 17.1* & 23.7 & \underline{26.0} & \tablelargesize\textbf{28.6} \\
    MMMU$_{\mathrm{val}}$ & 38.8 & \underline{43.6} & 41.1 & \tablelargesize\textbf{45.3} \\
    BLINK$_{\mathrm{test}}$ & 42.3* & \underline{44.0} & \nt43.1$^*$ & \tablelargesize\textbf{44.2} \\

    % \hline
    % \hline
    \rowcolor[gray]{0.9}
    \multicolumn{5}{l}{\emph{Knowledge/General QA}} \\

    RealWorldQA & 48.8* & 60.1 & \underline{62.9} & \tablelargesize\textbf{67.3} \\
    AI2D & 62.1* & \underline{74.9} & 69.9 & \tablelargesize\textbf{78.2} \\
    GQA & 49.2* & 59.5* & \nt\underline{59.8}$^*$ & \tablelargesize\textbf{62.7} \\
    MME & 1600* & \nt\np\textbf{2005}$^*$ & \np1872 & \tablelargesize\np\underline{1901} \\
    % MM-Vet & - & 60.8 & 49.5 & 51.0 \\
    \thickhline
  \end{tabularx}
  \label{tab:2bimageeval}
\end{table*}

\begin{table*}[t]
  \tablesize
  \centering
  \renewcommand{\arraystretch}{1.4}
  \caption{\textbf{Evaluation results of 7B models on image benchmarks.} $^*$ denotes the reproduced results. $^{\dagger}$ denotes the results retrieved from the official leaderboard. The best results are \textbf{in bold} and the second best ones are \underline{underlined}.}
  \begin{tabularx}{\textwidth}{ p{3.0cm} | >{\centering\arraybackslash}X | >{\centering\arraybackslash}X | >{\centering\arraybackslash}X | >{\centering\arraybackslash}X | >{\centering\arraybackslash}X | >{\centering\arraybackslash}X }
    \thickhline

    % \diagbox[width=3.2cm]{\textbf{Benchmark}}{\textbf{Model}} & 
    % \multirow{2}{*}{} & \multicolumn{2}{c|}{Open-weight Models} & \multicolumn{5}{c}{Open-source Models} \\
    % \cline{2-8}
    &
    \rotatebox{60}{\makecell[l]{\textbf{Molmo-7B-D} \\ \textbf{7B} \ \includegraphics[height=6pt]{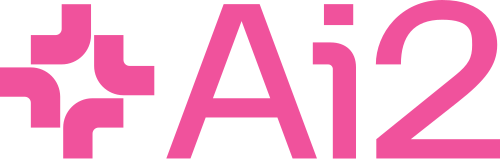}}} & 
    \rotatebox{60}{\makecell[l]{\textbf{InternVL2.5} \\ \textbf{8B}\ \ \includegraphics[height=8.5pt]{figures/icons/opengvlab.jpeg}}} & 
    \rotatebox{60}{\makecell[l]{\textbf{LLaVA-OneVision}\ \ \\ \textbf{7B}\ \ \includegraphics[height=8pt]{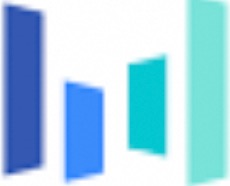}}} & 
    \rotatebox{60}{\makecell[l]{\textbf{NVILA} \\ \textbf{8B}\ \ \includegraphics[height=8pt]{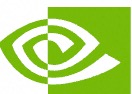}}} & 
    \rotatebox{60}{\makecell[l]{\textbf{Qwen2-VL} \\ \textbf{7B}\ \ \includegraphics[height=8pt]{figures/icons/qwen.png}}} & 
    \rotatebox{60}{\makecell[l]{\textbf{VideoLLaMA3} \\ \textbf{7B}}} \\

    % \cline{3-10}

    \thickhline

    \rowcolor[gray]{0.9}
    \multicolumn{7}{l}{\emph{Document/Chart/Scene Text Understanding}} \\
    ChartQA                  & 84.1 & 84.8 & 80.0 & \underline{86.1} & 83.0 & \tablelargesize\textbf{86.3} \\
    DocVQA$_{\mathrm{test}}$  & 92.2 & 93.0 & 87.5 & 93.7 & \underline{94.5} & \tablelargesize\textbf{94.9} \\
    InfoVQA$_{\mathrm{test}}$ & 72.6 & \underline{77.6}   & 68.8 & 70.7 & 76.5 & \tablelargesize\textbf{78.9} \\
    OCRBench                  & -    & 822\np  & 621\np  & 676*    & \textbf{845}\np & \tablelargesize\underline{828}\phantom{.} \\

    % \hline
    % \hline
    \rowcolor[gray]{0.9}
    \multicolumn{7}{l}{\emph{Math}} \\

    MathVista$_{\mathrm{testmini}}$ & 51.6 & 64.4 & 63.2 & \underline{65.4} & 58.2 & \tablelargesize\textbf{67.1} \\
    MathVision$_{\mathrm{test}}$    & -    & \underline{19.7} & - & 11.9* & 16.3 & \tablelargesize\textbf{26.2} \\

    % \hline
    % \hline
    \rowcolor[gray]{0.9}
    \multicolumn{7}{l}{\emph{Multi Image}} \\

    MMMU-Pro & - & \tablelargesize\textbf{34.3} & \nt24.1$^{\dagger}$ & \nt29.5* & \nt31.4$^*$ & \tablelargesize\underline{33.6} \\
    MMMU$_{\mathrm{val}}$ & 45.3 & \textbf{56.0} & 48.8 & \underline{49.9} & 54.1 & \tablelargesize{48.8} \\
    BLINK$_{\mathrm{test}}$ & - & \underline{54.8} & 48.2 & \nt47.0$^*$ & \nt43.1$^*$ & \tablelargesize\textbf{56.7} \\

    % \hline
    % \hline
    \rowcolor[gray]{0.9}
    \multicolumn{7}{l}{\emph{Knowledge/General QA}} \\

    RealWorldQA & \underline{70.7} & 70.1 & 66.3 & 68.6 & 70.1 & \tablelargesize\textbf{72.7} \\
    AI2D & \textbf{93.2} & 84.5 & 81.4 & \underline{92.3} & 83.0 & \tablelargesize{84.7} \\
    GQA & - & - & \underline{62.3} & - & \nt62.4$^*$ & \tablelargesize\textbf{64.9} \\
    MME & - & \np\textbf{2344} & \np1998 & \np2219 & \np\underline{2327} & \tablelargesize{\np2102} \\
    % MM-Vet & - & \nt\textbf{62.8}$^{\dagger}$ & \nt57.5$^{\dagger}$ & - & 62.0$^{\dagger}$ & \tablelargesize\underline{62.3} \\
    \thickhline
  \end{tabularx}
  
  \label{tab:7bimageeval}
\end{table*}

\subsection{Image-based Evaluation}
\subsubsection{Baselines}
To comprehensively evaluate the image performance of VideoLLaMA3, we compare it against a diverse set of baselines. For the 2B version of the model, we select several strong methods, including SmolVLM~\citep{smolvlm2023}, InternVL2.5-2B~\citep{internvl2.5}, and Qwen2VL-2B~\citep{wang2024qwen2}. For the 7B model, there are more options available. We choose to compare against Molmo-7B-D~\citep{molmo2024}, InternVL2.5-8B~\citep{internvl2.5}, LLaVA-OneVision~\citep{li2024llavaonevision}, NVILA~\citep{liu2024nvila}, and Qwen2VL-8B~\citep{wang2024qwen2}.

\subsubsection{Benchmarks}
To evaluate the image recognition and perception capabilities of VideoLLaMA3, we conduct assessments on several representative benchmarks commonly used in Image-LLMs. These benchmarks cover four dimensions: document/chart/scene text understanding, mathematical reasoning, multi-image understanding, and general knowledge QA.

\paragraph{Document/Chart/Scene Text Understanding.}
To evaluate VideoLLaMA3's ability to understand various forms of texts in images, including documents, charts, and scene text, we conduct assessments on a range of benchmarks. Specifically, we use: 1) DocVQA~\citep{mathew2021docvqa} for document understanding, which evaluates the model's ability to process and extract information from text in documents; 2) ChartQA~\citep{masry2022chartqa} and InfoVQA~\citep{mathew2021docvqa} for chart understanding, assessing the model's ability to interpret and reason about data presented in graphical forms such as bar charts and line graphs; and 3) OCRBench~\citep{liu2023hidden} for scene text image understanding, which tests the model's capacity to extract and comprehend text from images of real-world scenes.

\paragraph{Mathematical Reasoning.}
VideoLLaMA3's mathematical reasoning capabilities are evaluated through the MathVista~\citep{lu2023mathvista} and MathVision~\citep{wang2024measuring} benchmarks. These benchmarks focus on evaluating the model's ability to reason about and solve mathematical problems presented in visual formats, including text-based mathematical expressions and problem-solving tasks that require visual interpretation.

\paragraph{Multi-image Understanding.}
To assess VideoLLaMA3's ability to understand and reason about multiple images in conjunction, we evaluate the model on several widely used benchmarks, including MMMU-Pro~\citep{yue2024mmmupro}, MMMU~\citep{yue2024mmmu}, and BLINK~\citep{fu2025blink}. These benchmarks test the model's ability to draw connections between images, handle multiple visual inputs.

\paragraph{General Knowledge QA.}
Finally, to evaluate VideoLLaMA3's performance in general question answering, particularly in real-world and complex scenarios, we conduct assessments using several challenging benchmarks. The benchmarks include: 1) RealWorldQA~\citep{realworldqa}, which focuses on answering questions based on realistic images drawn from everyday scenarios, 2) AI2D~\citep{kembhavi2016diagram}, which evaluates the model's ability to reason about diagrams and science images, 3) GQA~\citep{hudson2019gqa}, which assesses general question answering with a focus on complex visual reasoning tasks, and 4) MME~\citep{fu2023mme}, which includes a wide variety of general knowledge questions that require a deep understanding of visual information.

\subsubsection{Evaluation Protocols}

When evaluating on benchmarks, we set the temparature as $0.0$. The maximum token length is set as the same as the training stage. For benchmarks involving the MCQ, we will give the prompt like ``Answer with the option letter from the given choices directly.''. For the benchmarks with short answers, we will give the prompt like ``Answer the question with a single word or phrase.''. We follow the original benchmarks to calculate the final scores, and we also align our evaluation protocols with other evaluation toolkits, such as lmms-eval~\cite{zhang2024lmmsevalrealitycheckevaluation, lmms_eval2024} and VLMEvalKit~\cite{duan2024vlmevalkit}.

\subsubsection{Evaluation Results}
We evaluate our VideoLLaMA3 model on the previously mentioned benchmarks. The evaluation results for our 2B model are presented in Table~\ref{tab:2bimageeval}. As shown, VideoLLaMA3 demonstrates significant improvements across a range of tasks compared to prior models. For example, in OCR benchmarks such as InfoVQA, VideoLLaMA3 achieves a performance score of 69.4\%, compared to the previous best score of 65.5\%. In mathematical reasoning tasks, such as MathVista, our 2B model scores 59.2\%, surpassing the state-of-the-art method by 7.9\%. For multi-image benchmarks like MMMU-Pro, VideoLLaMA3 outperforms the previous top-performing method by 2.6\%. In real-world knowledge QA tasks, such as RealWorldQA, VideoLLaMA3 achieves the highest performance with a score of 67.3\%, compared to 62.9\% from prior methods.

Similarly, we evaluate our larger 7B model on various image benchmarks, with results summarized in Table~\ref{tab:7bimageeval}. From the table, it is clear that VideoLLaMA3 consistently outperforms prior models on most benchmarks. Notably, in mathematical reasoning tasks, our 7B model surpasses the previous best by 6.5\% on MathVision. In chart understanding tasks, we observe a 1.3\% performance improvement over previous methods on InfoVQA. Additionally, in general reasoning tasks like RealWorldQA, VideoLLaMA3 outperforms prior models by 2.0\%.

Overall, the results confirm that VideoLLaMA3 provides consistent advancements across a broad range of benchmarks, demonstrating its efficacy and versatility in handling complex tasks, including OCR, mathematical reasoning, and general knowledge. These improvements position VideoLLaMA3 as a powerful tool for real-world applications, advancing the field of multi-modal learning.

\begin{table}[t]
  \tablesize
  \centering
  \renewcommand{\arraystretch}{1.4}
  \caption{\textbf{Evaluation results of 2B models on video benchmarks.} * denotes the reproduced results. $^{\dagger}$ denotes the results retrieved from the official leaderboard. The best results are \textbf{in bold} and the second best ones are \underline{underlined}.}
  \begin{tabularx}{\textwidth}{ p{3.0cm} | >{\centering\arraybackslash}X | >{\centering\arraybackslash}X | >{\centering\arraybackslash}X | >{\centering\arraybackslash}X }
    \thickhline

    \diagbox[width=3.4cm]{\textbf{Benchmark}}{\textbf{Model}} & 
    \makecell{\includegraphics[height=0.3cm]{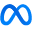} \textbf{Apollo} \\ \textbf{2B}} & 
    \makecell{\includegraphics[height=0.3cm]{figures/icons/opengvlab.jpeg} \textbf{InternVL2.5} \\ \textbf{2B}} & 
    \makecell{\includegraphics[height=0.3cm]{figures/icons/qwen.png} \textbf{Qwen2-VL} \\ \textbf{2B}} & 
    \makecell{\textbf{VideoLLaMA3} \\ 2B} \\
    % \cline{3-10}

    \thickhline
    \rowcolor[gray]{0.9}
    \multicolumn{5}{l}{\emph{General Video Understanding}} \\
    VideoMME \emph{\scriptsize w/o sub} & 53.0  & 51.9  & \underline{55.6}  & \tablelargesize\textbf{59.6} \\
    VideoMME \emph{\scriptsize w/ sub}  & 54.6  & 54.1  & \underline{60.4}  & \tablelargesize\textbf{63.4} \\
    MMVU$_{\mathrm{val}}$               & -     & \nt33.6$^*$ & \nt\underline{36.5}$^{\dagger}$ & \tablelargesize{\textbf{39.9}} \\
    MVBench                             & -     & \textbf{68.8}  & 63.2  & \tablelargesize\underline{65.5} \\
    EgoSchema$_{\mathrm{test}}$         & -     & \nt\underline{58.1}$^*$ & 54.9  & \tablelargesize\textbf{58.5} \\
    PerceptionTest$_{\mathrm{test}}$    & 61.0  & \nt\underline{66.3}$^*$ & 53.9  & \tablelargesize\textbf{68.0} \\
    ActivityNet-QA                      & -     & \nt\underline{54.1}$^*$ & \nt53.3$^*$ & \tablelargesize\textbf{58.2} \\

    % \hline
    % \hline
    \rowcolor[gray]{0.9}
    \multicolumn{5}{l}{\emph{Long Video Understanding}} \\

    MLVU$_{\mathrm{dev}}$               & \underline{63.3} & \nt58.9$^*$ & \nt62.7$^*$ & \tablelargesize\textbf{65.4} \\
    LongVideoBench$_{\mathrm{val}}$     & - & \underline{52.0} & \nt48.7$^*$ & \tablelargesize\textbf{57.1} \\
    % LVBench                             & -     & \nt37.3$^*$ & \nt\underline{38.0}$^*$ & \tablelargesize\textbf{40.4} \\
    LVBench                             & -     & \nt37.9$^*$ & \nt\underline{39.4}$^*$ & \tablelargesize\textbf{41.6} \\

    % \hline
    % \hline
    \rowcolor[gray]{0.9}
    \multicolumn{5}{l}{\emph{Temporal Reasoning}} \\

    TempCompass                         & 60.8 & \nt57.7$^*$ & \nt\underline{62.2}$^*$ & \tablelargesize\textbf{63.4} \\
    NextQA                              & -    & \nt75.6$^*$ & \nt\underline{77.2}$^*$ & \tablelargesize\textbf{81.1} \\

    % \hline
    % \hline
    % \rowcolor[gray]{0.9}
    % \multicolumn{5}{l}{\emph{Temporal Grounding}} \\

    Charades-STA                        & - & - & - & \tablelargesize\textbf{55.5} \\
    % ActivityNet                       & - & - & - & - \\

    \thickhline
  \end{tabularx}
  \label{tab:2bvideoresults}
  \vspace{1pt}
\end{table}

\begin{table}[t]
  \tablesize
  \centering
  \renewcommand{\arraystretch}{1.4}
  \caption{\textbf{Evaluation results of 7B models on video benchmarks.} * denotes the reproduced results. $^{\dagger}$ denotes the results retrieved from the official leaderboard. The best results are \textbf{in bold} and the second best ones are \underline{underlined}.}
  \begin{tabularx}{\textwidth}{ p{3.0cm} | >{\centering\arraybackslash}X | >{\centering\arraybackslash}X | >{\centering\arraybackslash}X | >{\centering\arraybackslash}X | >{\centering\arraybackslash}X | >{\centering\arraybackslash}X | >{\centering\arraybackslash}X }
    \thickhline

    % \diagbox[width=3.2cm]{\textbf{Benchmark}}{\textbf{Model}} & 
    % \multirow{2}{*}{} & \multicolumn{2}{c|}{Open-weight Models} & \multicolumn{5}{c}{Open-source Models} \\
    % \cline{2-8}
    &
    \rotatebox{60}{\makecell[l]{\textbf{Qwen2-VL} \\ \textbf{7B}\ \ \includegraphics[height=8pt]{figures/icons/qwen.png}}} & 
    \rotatebox{60}{\makecell[l]{\textbf{InternVL2.5} \\ \textbf{8B}\ \ \includegraphics[height=8.5pt]{figures/icons/opengvlab.jpeg}}} & 
    \rotatebox{60}{\makecell[l]{\textbf{LLaVA-Video}\ \ \\ \textbf{7B}\ \ \includegraphics[height=8pt]{figures/icons/bytedance.jpg}}} & 
    \rotatebox{60}{\makecell[l]{\textbf{NVILA} \\ \textbf{8B}\ \ \includegraphics[height=8pt]{figures/icons/nvidia.jpg}}} & 
    \rotatebox{60}{\makecell[l]{\textbf{Apollo} \\ \textbf{7B}\ \ \includegraphics[height=8pt]{figures/icons/meta.png}}} & 
    \rotatebox{60}{\makecell[l]{\textbf{VideoLLaMA} \\ \textbf{2.1-7B}}} & 
    \rotatebox{60}{\makecell[l]{\textbf{VideoLLaMA} \\ \textbf{3-7B}}} \\

    \thickhline
    \rowcolor[gray]{0.9}
    \multicolumn{8}{l}{\emph{General Video Understanding}} \\
    VideoMME \emph{\scriptsize w/o sub} & 63.3          & \underline{64.2}        & 63.3  & \underline{64.2}  & 61.3  & 54.9  & \tablelargesize\textbf{66.2} \\
    VideoMME \emph{\scriptsize w/ sub}  & 69.0          & 66.9                    & 69.7  & \underline{70.0}  & 63.3  & 56.4 & \tablelargesize\textbf{70.3} \\
    MMVU$_{\mathrm{val}}$               & \nt42.1$^{\dagger}$  & \nt41.1$^{\dagger}$ & \nt42.4$^*$ & \nt\underline{43.7}$^*$ & - & \nt39.5$^{\dagger}$ & \tablelargesize \textbf{44.1} \\
    MVBench                             & 67.0          & \textbf{72.0}           & 58.6  & 68.1  & -     & 57.3 & \tablelargesize\underline{69.7} \\
    EgoSchema$_{\mathrm{test}}$         & \textbf{66.7} & \nt\underline{66.2}$^*$ & 57.3  & \nt54.3$^*$       & -     & 53.1 & \tablelargesize{63.3} \\
    PerceptionTest$_{\mathrm{test}}$    & 62.3          & \nt68.9$^*$             & \nt\underline{67.9}$^*$   & \nt65.4$^*$ & -     & 54.9  & \tablelargesize\textbf{72.8} \\
    ActivityNet-QA                      & \nt57.4$^*$   & \nt58.9$^*$             & 56.5  & \underline{60.9}  & -     & 53.0  & \tablelargesize\textbf{61.3} \\

    % \hline
    % \hline
    \rowcolor[gray]{0.9}
    \multicolumn{8}{l}{\emph{Long Video Understanding}} \\

    MLVU$_{\mathrm{dev}}$               & \nt69.8$^*$ & \nt69.0$^*$ & \nt70.8$^*$ & \nt70.6$^*$ & \underline{70.9}  & 57.4  & \tablelargesize\textbf{73.0} \\
    LongVideoBench$_{\mathrm{val}}$     & \nt55.6$^{\dagger}$  & \textbf{60.0}  & 58.2  & 57.7  & 58.5  & -     & \tablelargesize\underline{59.8} \\
    % LVBench                             & \nt\textbf{44.2}$^*$ & \nt41.5$^*$ & \nt40.3$^*$ & \nt42.6$^*$ & -     & 36.3  & \tablelargesize\underline{43.7} \\
    LVBench                             & \nt\underline{44.7}$^*$ & \nt43.2$^*$ & \nt41.5$^*$ & \nt44.0$^*$ & -     & 36.2  & \tablelargesize\textbf{45.3} \\

    % \hline
    % \hline
    \rowcolor[gray]{0.9}
    \multicolumn{8}{l}{\emph{Temporal Reasoning}} \\

    TempCompass                         & \nt67.9$^{\dagger}$  & \nt\underline{68.3}$^*$ & 65.4  & \nt\textbf{69.7}$^*$ & 64.9 & 56.8  & \tablelargesize{68.1} \\
    NextQA                              & \nt81.2$^*$          & \nt\textbf{85.0}$^*$ & 83.2  & 82.2  & -    & 75.6  & \tablelargesize\underline{84.5} \\

    % \hline
    % \hline
    % \rowcolor[gray]{0.9}
    % \multicolumn{8}{l}{\emph{Temporal Grounding}} \\

    Charades-STA                        & -     & -     & -     & -     & -    & -     & \tablelargesize\textbf{60.7} \\
    % ActivityNet &  \\

    \thickhline
  \end{tabularx}
  \vspace{1pt}
  \label{tab:7bvideoresult}
\end{table}

\subsection{Video-based Evaluation}

\subsubsection{Baselines}
To comprehensively evaluate the video performance of VideoLLaMA3, we compare it with a diverse set of baseline models. Similar to image evaluation, there are few available models with a 2B parameter size in the community. We select several strong baselines, including Apollo-2B~\citep{zohar2024apollo}, InternVL2.5-2B~\citep{internvl2.5}, and Qwen2VL-2B~\citep{wang2024qwen2}. For the 7B model, we compare it with generalist models such as Qwen2VL-7B~\citep{wang2024qwen2}, InternVL2.5-8B~\citep{internvl2.5}, and NVILA~\citep{liu2024nvila}, as well as specialist models like LLaVA-Video~\citep{zhang2024video}, Apollo-7B~\citep{zohar2024apollo}, and our previous generation model, VideoLLaMA2~\citep{damonlpsg2024videollama2}.

\subsubsection{Benchmarks}
The video understanding capabilities of VideoLLaMA3 are systematically evaluated across three core dimensions: general understanding, temporal reasoning, and long-form video comprehension.

\paragraph{General Video Understanding.} We assess VideoLLaMA3's general video understanding capabilities through established benchmarks: (1) Multi-Choice Video Question Answering (MC-VQA) tasks, including MVBench~\citep{li2023mvbench}, VideoMME~\citep{fu2024video}, EgoSchema~\citep{mangalam2024egoschema}, and Perception-Test~\citep{patraucean2024perception}. (2) Open-Ended Video Question Answering (OE-VQA) tasks, including ActivityNet-QA~\citep{yu2019activitynet} and VCGBench~\citep{Maaz2023VideoChatGPT}. 
% These benchmarks involve multiple-choice questions with accuracy metrics, except ActivityNet-QA utilizes GPT-3.5 for answer quality assessment. 
This evaluation suite follows the protocol of VideoLLaMA2~\citep{damonlpsg2024videollama2}. We also run evaluations on MMVU~\citep{zhao2025mmvu} which includes both the task types mentioned above. 

\paragraph{Long Video Understanding.} To further examine the capacity of VideoLLaMA3 to process and comprehend long-form video content, we assess performance on three long-video understanding (LVU) benchmarks: (1) MLVU~\citep{MLVU}: diverse long-video understanding tasks for videos ranging from 3 minutes to more than 2 hours, 
(2) LongVideoBench~\citep{wu2024longvideobench}: video reasoning over the referred context within long video-language interleaved inputs,  and (3) LVBench~\citep{wang2024lvbench}: extreme long video understanding.

\paragraph{Video Temporal Reasoning.}
To assess the temporal awareness and reasoning capabilities of VideoLLaMA3, we conduct evaluations on the following tasks: (1) Temporal Perception and Reasoning tasks, including TempCompass~\citep{liu2024tempcompass} and NextQA~\citep{xiao2021next}; and (2) Temporal Sentence Grounding task on Charades-STA~\citep{gao2017tall} benchmark, with mean Intersection over Union (mIoU) metric.

\subsubsection{Evaluation Protocols}
% \lx{Note: I think we can give more details of evaluation here (like the maximum of video frames and sequence length ... )}

We expand the max number of visual tokens to 16K when evaluating our models on video-based benchmarks, ensuring that each frame corresponds to a reasonable number of tokens and the total context length is within the maximum range of the base LLM.
The maximum number of frames is set to 180, which is the same as training.
For reproducibility, we keep these hyperparameters the same on all benchmarks and disable sampling when decoding.

For general multi-chocies question answering evaluation, we follow the official setting to construct the instruction using provided questions and options.
An addition prompt like ``Answer with the option's letter from the given choices directly" is added to control the model output.
In addition, we apply CoT prompt on MMVU benchmark following the official evaluation protocol.
For temporal grounding evaluation, we add an extra prompt ``Please output the start and end timestamps in seconds" after the question.
The numbers in the model response are extracted by regular expression, and then treated as one or multiple time intervals.
Based on this strategy, we finally report the mIoU bewteen the ground-truth intervals and the predicted intervals.

\subsubsection{Evaluation Restuls.}
Table~\ref{tab:2bvideoresults} evaluates the performance of Video Understanding models with 2B model size. VideoLLAMA3 consistently demonstrates competitive results and outperforms baseline methods. In General Video Understanding, VideoLLAMA3 achieves the highest scores on VideoMME w/o sub (59.6\%), VideoMME w/ sub (63.4\%), ActivityNet-QA (58.2\%),  PerceptionTest-test (68.0\%), MVBench (65.5\%), and MMVU (37.6\%). On MVBench, it ranks second (65.5\%), slightly behind InternVL2.5 2B (68,8\%).
For Long Video Understanding, VideoLLAMA3 achieves the best performance on all benchmarks: MLVU-dev (65.4\%), LongVideoBench-val (57.1\%), and LVBench (40.4\%), showcasing its superior ability to handle long video content.
In Temporal Reasoning, VideoLLAMA3 leads on TempCompass (63.4\%), and NextQA (81.1\%), and Charades-STA (55.5\%).
Compared to Apollo-2B, InternVL2.5-2B, and Qwen2-VL-2B, VideoLLAMA3 not only secures the top position in most benchmarks but also demonstrates consistent superiority in tasks requiring comprehensive and long-term video understanding, reinforcing its strong capability across diverse video-related tasks.

As for the VideoLLaMA3-7B model, the results are shown in Table~\ref{tab:7bvideoresult}.
On 7B model size, VideoLLaMA3-7B still exhibits competitive results. For general video understanding, it leads on 5 out of 7 benchmarks, including VideoMME w/o sub, VideoMME w/ sub, PerceptionTest-test, and ActivityNet-QA. On MVBench, it also achieves comparable results to InternVL2.5-8B. For long video understanding, VideoLLaMA3-7B scores the highest on MLVU-dev, and achieves the second best results on LongVideoBench-val and LVBench.

\subsection{Case Study}

\begin{figure*}[tbp] 
\centering
\includegraphics[width=1.\linewidth]{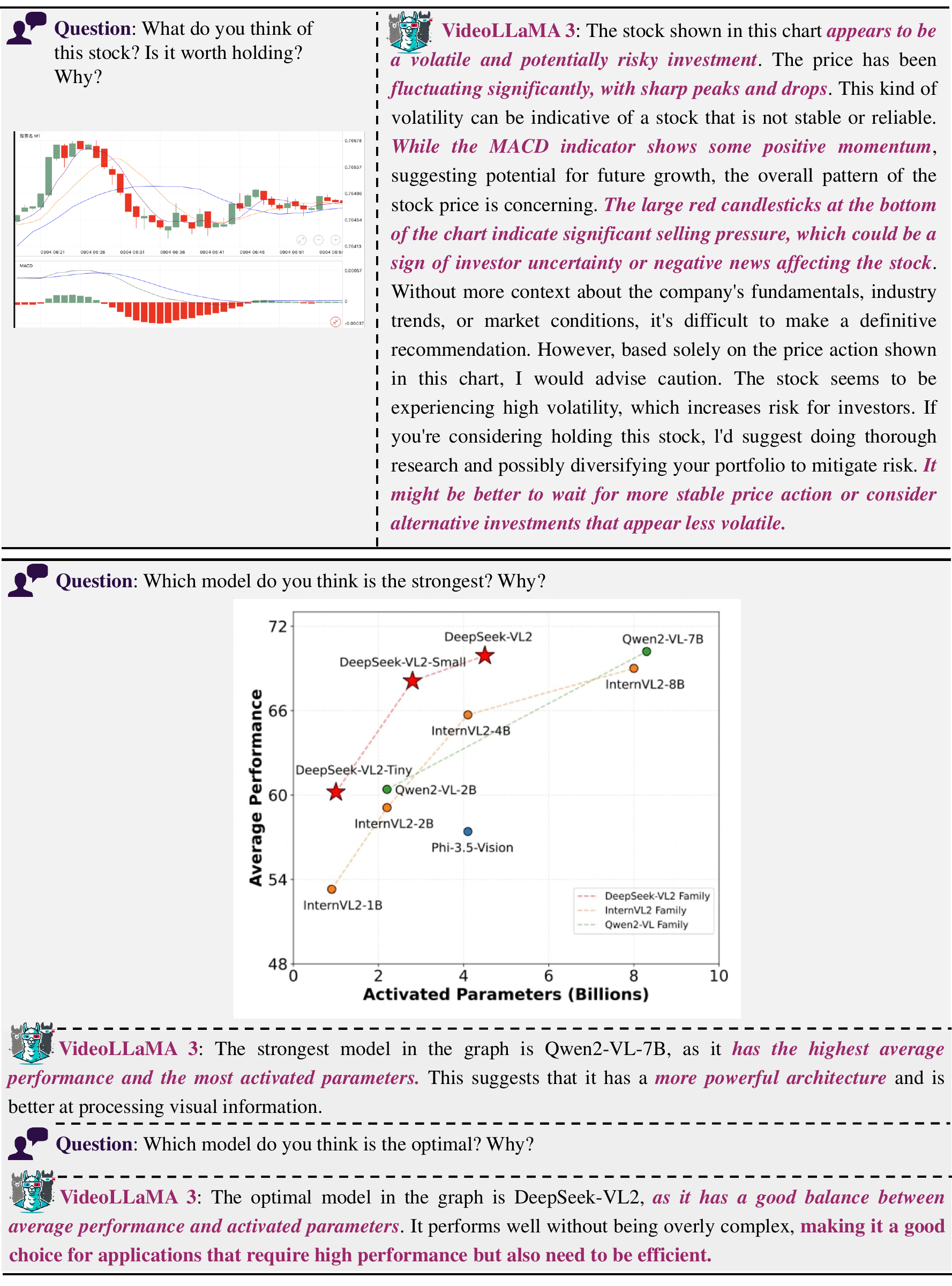} 
\caption{Case study of chart images understanding.
}
\label{fig:case_chart}
\end{figure*}

\noindent\textbf{Chart Image Understanding.} In Figure~\ref{fig:case_chart}, we show two cases for chart image understanding. In the first case, VideoLLaMA3 can analyze stock trends and offer some reasonable suggestions for investment. As for the second case, the model can compare the performance of MLLMs and know the tradeoff between the number of paramters and performances.

\begin{figure*}[tbp] 
\centering
\includegraphics[width=1.\linewidth]{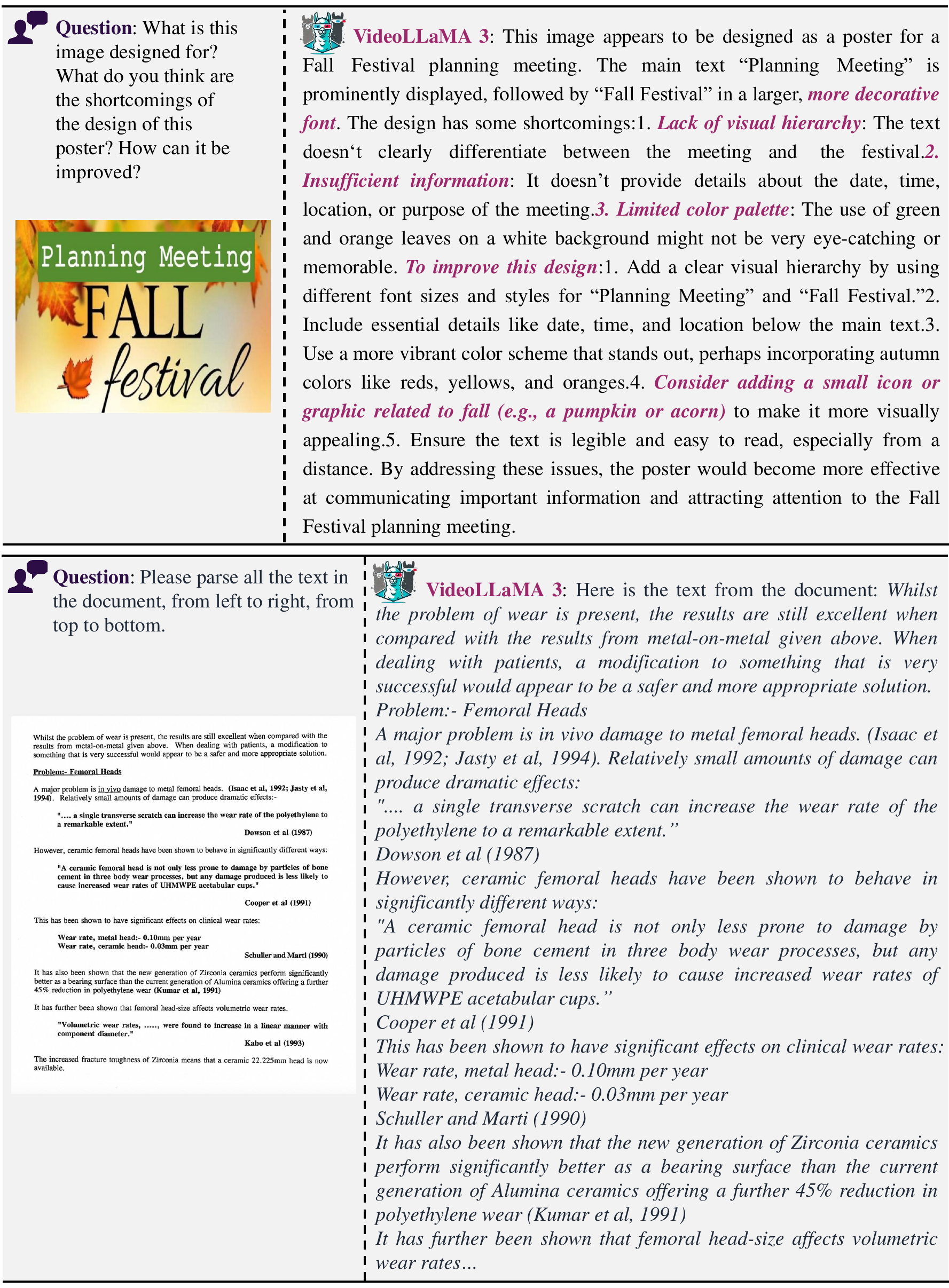} 
\caption{Case study of OCR and document images.
}
\label{fig:case_ocr}
\end{figure*}

\noindent\textbf{OCR and Document Understanding.} Figure~\ref{fig:case_ocr} shows two cases for images with texts. In this first example, the model can successfully parse the words in the design image, and offer some suggestions to make the poster better. In the second image, we ask VideoLLaMA3 to perform OCR task on the given document image. VideoLLaMA3 can successfully recognize the words in the document image, demonstrating the strong performance of VideoLLaMA3 in understanding dense information in images.

\begin{figure*}[tbp] 
\centering
\includegraphics[width=1.\linewidth]{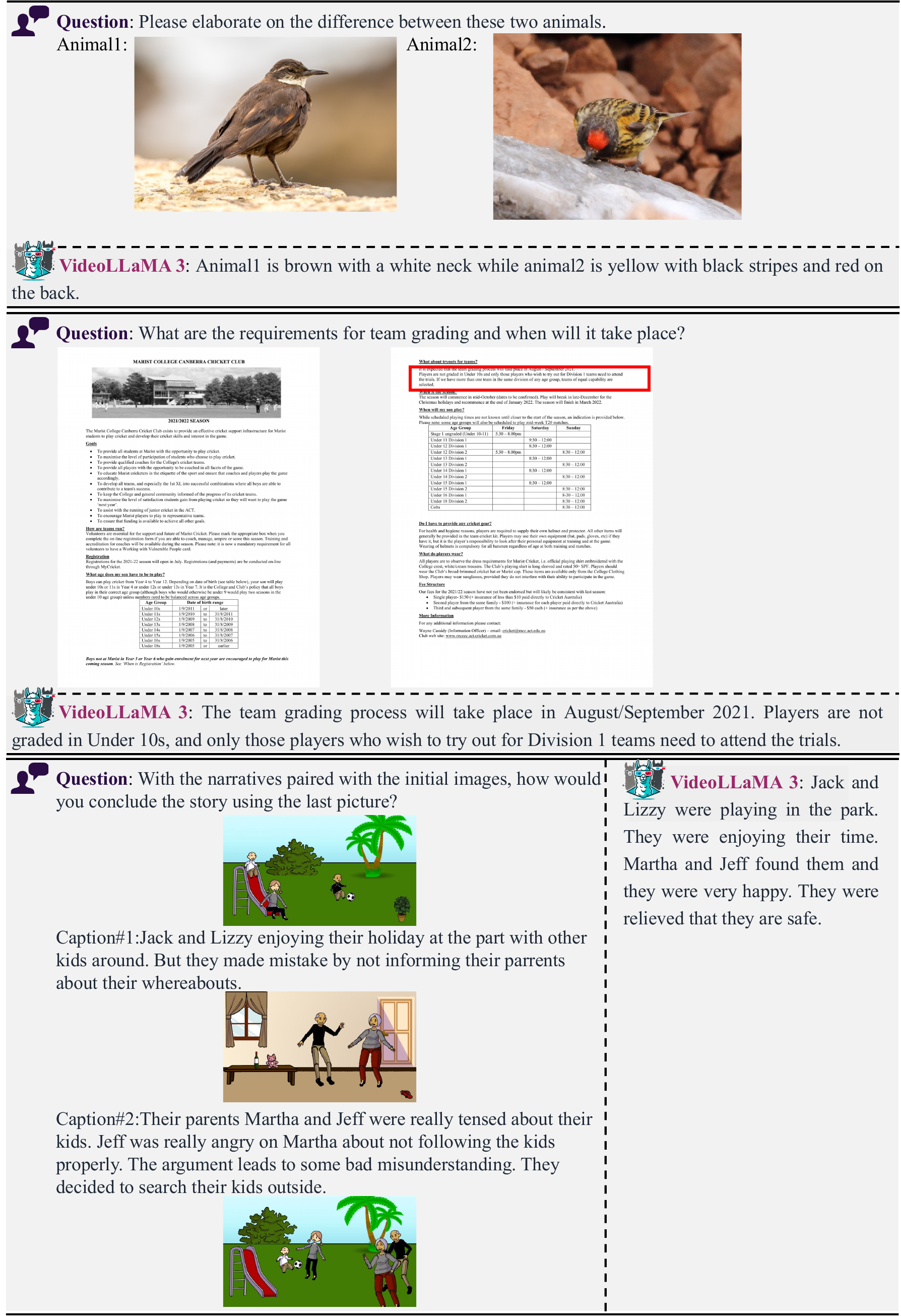} 
\caption{Case study of multi-image understanding.
}
\label{fig:case_multi_image}
\end{figure*}

\noindent\textbf{Multi-Image Understanding.} Figure~\ref{fig:case_multi_image} gives three examples on multi-image understanding tasks. In the first example, VideoLLaMA3 can tell the differences between two types of birds. The second example demonstrates that VideoLLaMA3 is able to locate answers from long documents (even with multiple images) rather than simplying parsing words. It is an advanced capability beyond OCR. While in the last example, VideoLLaMA3 can understand storylines from comic strips.

\begin{figure*}[tbp] 
\centering
\includegraphics[width=1.\linewidth]{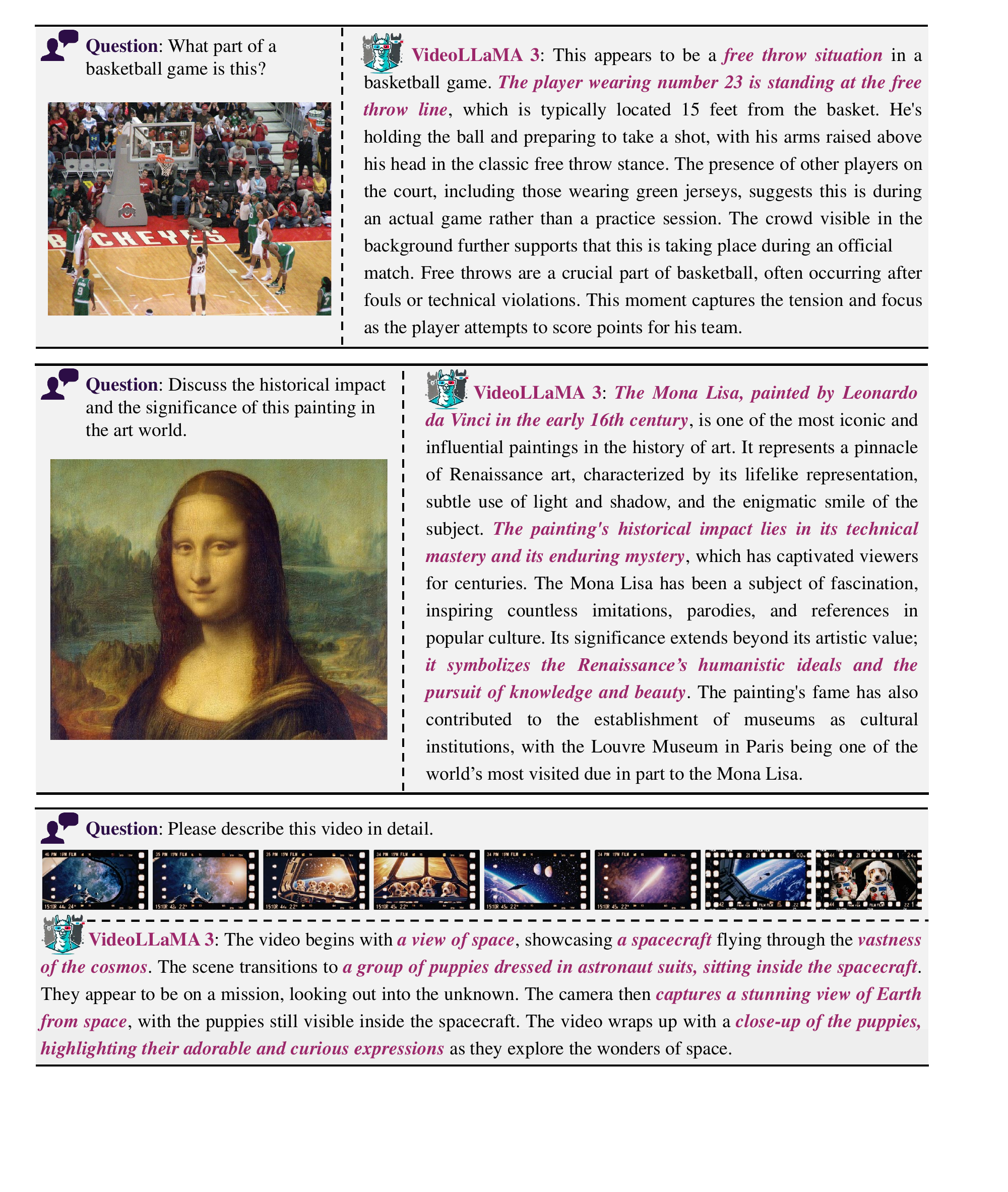} 
\caption{Case study of images with general knowledge.
}
\label{fig:case_general}
\end{figure*}

\begin{figure*}[tbp] 
\centering
\includegraphics[width=1.\linewidth]{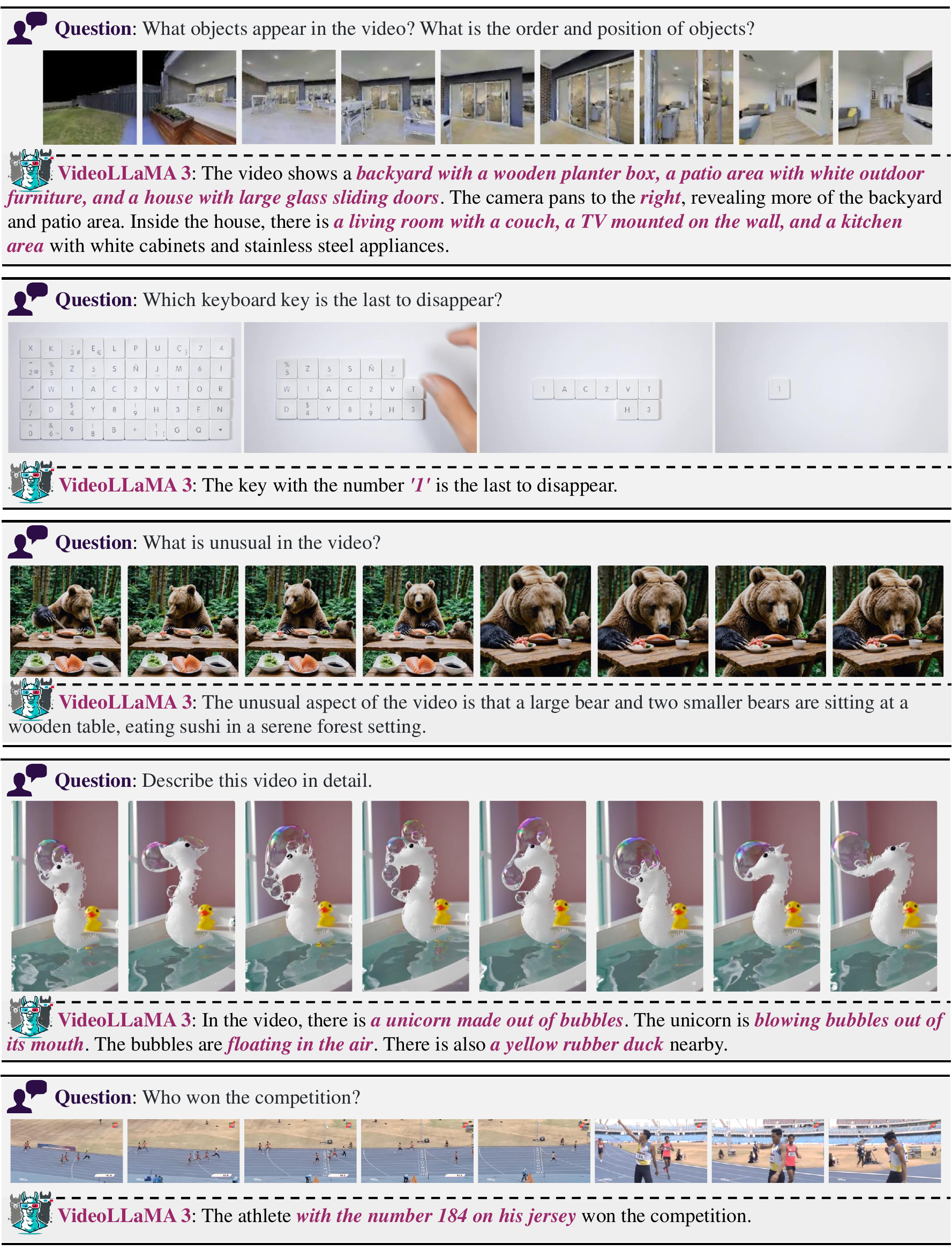} 
\caption{Case study of video understanding.
}
\label{fig:case_video}
\end{figure*}

\noindent\textbf{General Image and Video Understanding.} Figure~\ref{fig:case_general} demonstrates VideoLLaMA3's capability in understanding general images, including VQA tasks, answering questions using knowledges and providing videos with captions. Also in Figure~\ref{fig:case_video}, we give fives cases for video understanding. VideoLLaMA3 can comprehend video content through temporal dimensions, rather than relying solely on inferences from static content.

\begin{figure*}[tbp] 
\centering
\includegraphics[width=1.\linewidth]{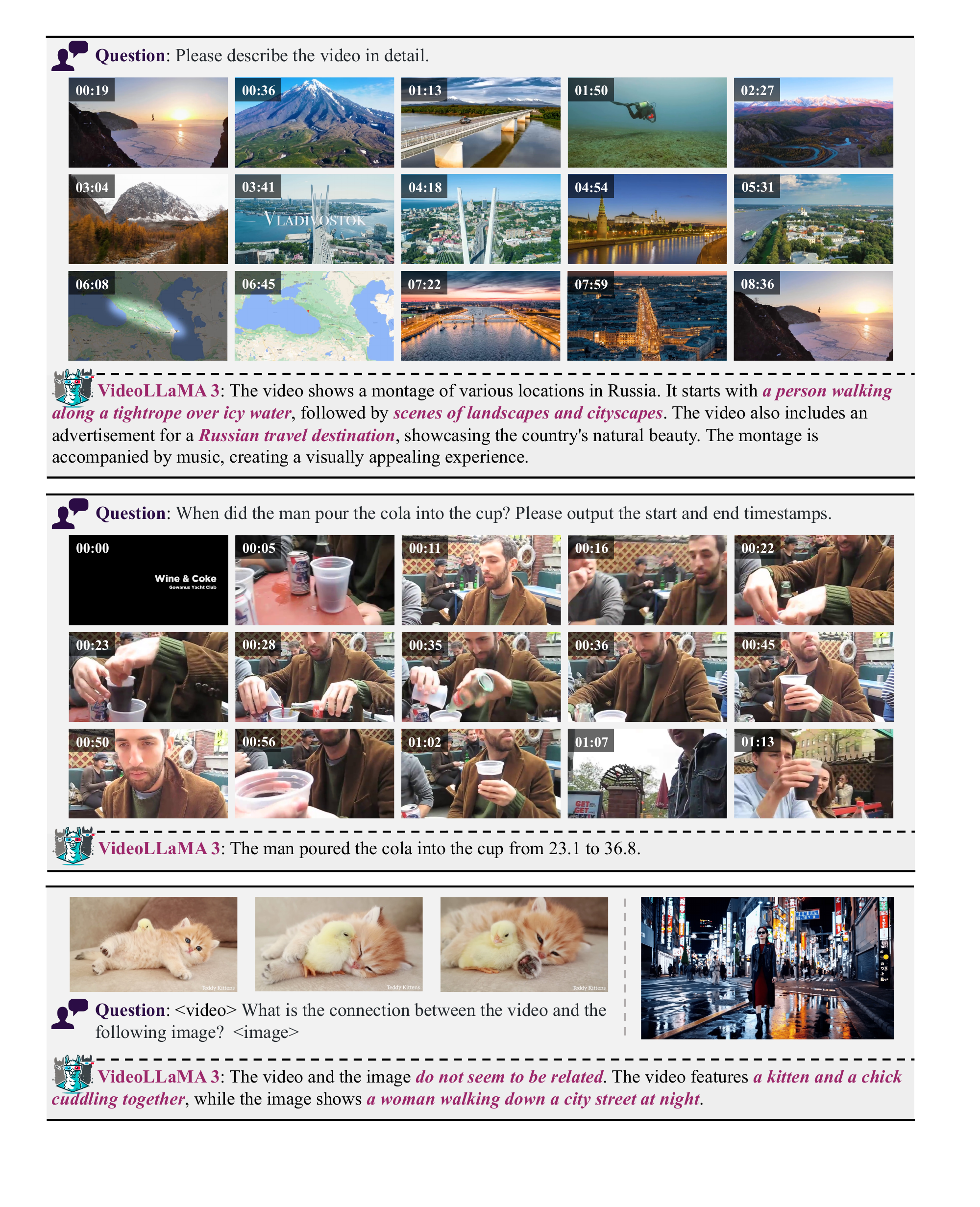} 
\caption{Case study of long video understanding, temporal grounding, and video-image joint understanding.
}
\label{fig:case_complex_video}
\end{figure*}

\noindent\textbf{Long video understanding, temporal grounding, and video-image joint understanding.} In Figure~\ref{fig:case_complex_video}, we present several cases involving more complex video tasks, including long video grounding, video temporal grounding, and video-image joint understanding. Our VideoLLaMA3 model demonstrates the ability to perform complex long video question-answering tasks. For tasks requiring temporal grounding, our model accurately identifies the specified time. Additionally, for video-image joint understanding, the model effectively captures the relationships between videos and images, enabling it to tackle more intricate tasks.

% \noindent\textbf{Video Captioning.} xxxx

% \noindent\textbf{Long Video Understanding.} xxxx

% \noindent\textbf{Temporal Grounding.} xxx

\subsection{Ablation Study}

\begin{table}[h]
    \centering
    % \tablesize
    % \renewcommand{\arraystretch}{1.4}

    % \begin{tabularx}{\textwidth}{ p{4.5cm} | >{\centering\arraybackslash}X | >{\centering\arraybackslash}X | >{\centering\arraybackslash}X | >{\centering\arraybackslash}X | >{\centering\arraybackslash}X | >{\centering\arraybackslash}X }
    %     \thickhline
    %     \textbf{Model}  & \textbf{GQA}  & \textbf{AI2D} & \textbf{ChartQA} & \textbf{DocVQA$_{\mathrm{val}}$} & \textbf{MME} & \textbf{TextVQA} \\
    %     \thickhline
    %     clip-vit-large-patch14-336~\cite{radford2021learning}   & 61.50 & 56.28 & 18.32 & 24.86 & \textbf{1668.41} & 52.81 \\
    %     dfn5B-clip-vit-h-14-378~\cite{fang2023data}  & 62.70 & 56.87 & 16.40 & 23.09 & 1665.35 & 52.24 \\ 
    %     siglip-so400m-patch14-384~\cite{Zhai2023SigmoidLF} & \textbf{62.92} & \textbf{57.12} & \textbf{22.44} & \textbf{31.32} & 1667.92 & \textbf{57.24} \\ 
    %     \thickhline
    % \end{tabularx}
    \caption{\textbf{Ablation Study on Vision Encoders.}}
    \resizebox{\textwidth}{!}{
    \renewcommand{\arraystretch}{1.5}
    \begin{tabular}{lccccc}
    \thickhline
    \textbf{Model}  & \textbf{GQA}  & \textbf{AI2D} & \textbf{ChartQA} & \textbf{DocVQA$_{\mathrm{val}}$} & \textbf{MME} \\ \thickhline
    clip-vit-large-patch14-336~\cite{radford2021learning}   & 61.50 & 56.28 & 18.32 & 24.86 & \textbf{1668.41} \\
    dfn5B-clip-vit-h-14-378~\cite{fang2023data}  & 62.70 & 56.87 & 16.40 & 23.09 & 1665.35 \\ 
    siglip-so400m-patch14-384~\cite{Zhai2023SigmoidLF} & \textbf{62.92} & \textbf{57.12} & \textbf{22.44} & \textbf{31.32} & 1667.92  \\ \thickhline
    \end{tabular}
    }
    \centering
    \label{tab:ablation}
\end{table}

% \noindent\textbf{The Choice of Vision Encoder.}
In MLLMs, the embeddings of pre-trained vision encoder should be trained to align with embeddings of LLMs. Therefore, the representation performance of vision encoder is crucial to the final performance of MLLMs.
In this work, we study the impact of different vision encoders. Specifically, we compare three pre-trained transformer-based vision encoders: CLIP~\cite{radford2021learning}, DFN~\cite{fang2023data}, and SigLIP~\cite{Zhai2023SigmoidLF}. Due to the computation limitation, we perform the study on the subset of the whole dataset. Also, to investigate the performance of the original pre-trained weights, we fix the weights of vision encoders and keep the visual inputs as the fixed resolution, which is the same as the pretrained resolution of the vision encoder ($336\times336$ for CLIP, $378\times378$ for DFN, and $384\times384$ for SigLIP,). The training has three stages: 1) Training projector with LLaVA-Pretrain-558K~\citep{liu2023improvedllava}; 2) Tuning all parameters with our recaptioned COYO data; 3) SFT with LLaVA-SFT-665K~\citep{li2024llava}.
% \begin{wraptable}{r}{0.5\textwidth} % r 表示右侧环绕，0.5\textwidth 表示宽度为半页
% \centering
% \begin{tabular}{lcc}
% \thickhline
% Model  & GQA  & TextVQA \\ \thickhline
% CLIP   & 61.50 & 52.81   \\
% DFN    & 62.70 & 52.24   \\ 
% SigLIP & 62.92 & 57.24   \\ \thickhline
% \end{tabular}
% \caption{Ablation Study on Vision Encoders.}
% \label{tab:ablation}
% \end{wraptable}
% \begin{wraptable}{r}{0.5\textwidth} % r 表示右侧环绕，0.5\textwidth 表示宽度为半页
% \centering
% \renewcommand{\arraystretch}{1.5}
% \begin{tabular}{lccccc}
% \thickhline
% \textbf{Model}  & \textbf{GQA}  & \textbf{AI2D} & \textbf{ChartQA} & \textbf{DocVQA$_{\mathrm{val}}$} & \textbf{MME} \\ \thickhline
% clip-vit-large-patch14-336   & 61.50 & 56.28 & 18.32 & 24.86 & \textbf{1668.41} \\
% dfn5B-clip-vit-h-14-378  & 62.70 & 56.87 & 16.40 & 23.09 & 1665.35\\ 
% siglip-so400m-patch14-384 & \textbf{62.92} & \textbf{57.12} & \textbf{22.44} & \textbf{31.32} & 1667.92 \\ \thickhline
% \end{tabular}
% \centering
% \caption{Ablation Study on Vision Encoders.}
% \label{tab:ablation}
% \end{wraptable}
The comparison results are shown in Table.~\ref{tab:ablation}. SigLIP outperforms the other two vision encoders, especially in fine-grained understanding tasks involving texts.
Based on this ablation study, we choose the pretrained SigLIP as our base vision encoder, and then adapt it to taking dynamic resolutions as inputs.

%% file: sections/related_work.tex
\section{Related Work}

\noindent\textbf{Multimodal LLMs for Native Video Understanding.} 
% from training recipe and data mixture perspective
% Streaming: Flash-VStream, VideoStreaming, VideoLLM-online, InternLM-XComposer2.5-OmniLive, StreamChat
% AV: VideoSalmon，VideoLLaMA, VideoLLaMA 2, CAT, AV-LLM, PandaGPT, MacawLLM, X-InstructBLIP, OneLLM, CREMA, AVicuna
Early video MLLMs primarily relied on sparsely sampled frames and simple connectors, such as MLPs~\citep{maaz2023video,lin2023video,ataallah2024minigpt4}, discrete visual tokenizers~\citep{jin2024video}, and Q-formers~\citep{zhang2023video,li2024mvbench}, to link visual encoders with large language models. 
Subsequent models propose various methods to overcome token limitations and support long-form video understanding. For example, ~\citep{zhang2024longva} directly extends the context window of LLMs to achieve long video understanding, while others~\citep{shen2024longvu,xu2024pllava,jin2024chat,zhang2024video,xu2024slowfast,weng2025longvlm,song2024moviellm,li2025llama,tan2024koala} introduce video token compression techniques that perform pooling across spatial, temporal, or both dimensions.
% To overcome token limitations and support long-form video understanding, subsequent models introduced spatio-temporal pooling for frame resampling~\citep{shen2024longvu,xu2024pllava,jin2024chat,zhang2024video,xu2024slowfast,weng2025longvlm,song2024moviellm,li2025llama,zhang2024longva,tan2024koala}. 
While most approaches utilize image-based encoders~\citep{zohar2024apollo,song2024moviechat+,chen2024sharegpt4video,fei2024video,wang2024tarsierrecipestrainingevaluating,hong2024cogvlm2,liu2024kangaroo,liu2024oryx,chen2024timemarker,wu2025valley2}, some incorporate video-specific encoders to better capture temporal dependencies~\citep{wang2024internvideo2,jung2024pegasus,Maaz2024VideoGPT+,cheng2024videollama}. 

More recent works~\citep{sun2024video,cheng2024videollama,ye2025cat,shu2023audio,su2023pandagpt,lyu2023macaw,panagopoulou2023x,han2024onellm,yu2024crema,tang2024avicuna} extend beyond visual inputs by incorporating audio, using separate encoders for each modality and integrating them through an LLM decoder. These models leverage joint instruction tuning on video-audio datasets~\citep{Yang2022AVQAAD,AlAmri2019AudioVS,ye2024cat,leng2024curse} to capture interactions between visual and auditory information. 
Additionally, recent advances in streaming video understanding focus on real-time processing~\citep{zhang2024flash,qian2024streaming,Chen2024VideoLLMonlineOV,Zhang2024InternLMXComposer25OmniLiveAC,Liu2024StreamChatCW}, employing techniques like adaptive memory and incremental processing for tasks such as live event detection and real-time captioning.
Previous works~\cite{shen2024longvu,zhang2024video,tan2024koala,zohar2024apollo,wang2024internvideo2,cheng2024videollama} typically follow a training recipe that involves an alignment phase, followed by supervised fine-tuning, with instruction-tuning datasets~\citep{zhang2024video,Maaz2023VideoChatGPT,li2023mvbench,liu2024oryx,chen2024sharegpt4video} often being video dominant. 
% The video instruction tuning datasets have evolved from small-scale to large-scale collections [cite], enabling more comprehensive training. 
However, we propose a vision-centric training paradigm to enhance video understanding capabilities by focusing on large-scale image understanding pre-training. This approach leverages high-quality image-text datasets to build robust vision encoders that are then adapted for video tasks.

\noindent\textbf{Multimodal LLMs for General Vision Understanding.}
Recently, a growing number of general MLLMs have been developed to process both images and videos. While, in principle, models designed to handle multiple images are inherently capable of processing video data, achieving optimal performance requires dedicated training on video-specific datasets.

Previous studies \cite{jiang2024mantisinterleavedmultiimageinstruction,li2024llavanextinterleavetacklingmultiimagevideo,li2024llavaonevision,mplug3} have demonstrated that general MLLMs with robust image understanding capabilities can achieve remarkable performance on video understanding tasks, even with minimal or no dedicated video training data. These findings highlight the effectiveness of task transfer from images to videos, showcasing the models' strong video comprehension and cross-scenario adaptability.

Furthermore, Qwen2-VL \cite{wang2024qwen2} adopts a unified framework for processing both images and videos, enhancing the model's visual perception capabilities. Models such as Qwen2-VL, InternVL-2 \cite{chen2023internvl}, and InternVL-2.5 \cite{internvl2.5}, which scale both model sizes (ranging from 1 billion to 78 billion parameters) and the volume of training data, have achieved highly competitive performance in both image and video understanding tasks. To address the challenges of processing longer video inputs, recent studies \cite{longllava,internlmxcomposer25} have proposed solutions such as adapting model architectures by incorporating a hybrid design of Mamba and Transformer blocks or training with extensive long video datasets to support extended input and output sequences.

% Information is available in diverse modalities. 
Recent studies \cite{vita,vita15,llama3} have integrated text, image, and video modalities with audio and speech modalities to improve models' video understanding and cross-scenario performance. Additionally, Aria \cite{aria} utilizes a fine-grained mixture-of-experts decoder, which enables more efficient training and inference compared to dense decoders when handling multimodal inputs.

% \subsection{Image MLLMs}

% xxx

% \subsection{Video MLLMs}
% xxx

%% file: sections/conclusion.tex
\section{Discussion, Limitations, and Future Work}

\subsection{Discussion}
The introduction of VideoLLaMA3 marks a significant advancement in the realm of MLLMs, particularly in bridging the gap between image and video understanding. By adopting a vision-centric training paradigm, VideoLLaMA3 leverages the robustness of image-centric data to enhance video comprehension, effectively mitigating the challenge associated with temporal dynamics and the complexity of video data. This approach underscores the inferent value of high-quality image-text datasets, which are more readily available and easier to curate compared to their video-text counterparts. The success of VideoLLaMA3 on diverse benchmarks, including VideoMME, PerceptionTest, MLVU, DocVQA, and MathVista, demonstrates its versatility and efficacy across various multimodal tasks.

The model's ability to maintain strong performance in both image and video domains highlights the effectiveness of our vision-centric framework designs. Specifically, the dynamic resolution adaptation and vision token compression strategies facilitate a more flexible and efficient representation of visual inputs, enabling the model to handle a wide range of image and video formats with minimal information loss. This flexibility is crucial for real-world applications where visual data can vary significantly in resolution and aspect ratio.

Furthermore, the multi-task fine-tuning stage of our training paradigm contributes to the model's robust generalization capabilities. By exposing VideoLLaMA3 to a variety of downstream tasks, including interactive question answering and video captioning, the model develops a comprehensive understanding of both static and dynamic visual information. This comprehensive training enables VideoLLaMA3 to excel not only in standard benchmarks but also in specialized tasks that require nuanced comprehension of visual content.

\subsection{Limitations}

Despite the impressive performance of VideoLLaMA3, several limitations must be acknowledged.

\noindent\textbf{Video Data Quality and Diversity.} While leveraging large-scale image-text datasets has proven beneficial, the quality and diversity of video-text datasets remain a constraint. Video data often suffer from lower annotation quality and limited diversity, which can impede the model's ability to generalize across different video domains and genres.

\noindent\textbf{Real-time Processing.} The current model architecture may not be optimized for real-time video processing tasks, which is essential for applications such as autonomous driving and live video analytics. The computational overhead associated with processing high-resolution and lengthy video inputs can hinder real-time performance.

\noindent\textbf{Generalization to Unseen Modalities.} While VideoLLaMA3 excels in image and video understanding, its capability to generalize to other modalities, such as audio or speech data, remains unexplored. Integrating additional modalities could further enhance the model's multimodal comprehension but poses significant challenges in terms of architecture and training.

\subsection{Future Work}

Building on the foundations laid by VideoLLaMA3, several avenues for future research are proposed to address the identified limitations and further enhance the model's capabilities.

\noindent\textbf{Enhanced Video-Text Datasets.} Investing in the creation and curation of higher quality and more diverse video-text datasets will be crucial. Incorporating annotations that capture nuanced temporal and contextual information can significantly improve the model's temporal understanding and generalization across different video domains.

\noindent\textbf{Real-time Inference Optimization.} Optimizing the model architecture for real-time inference by reducing latency and improving processing speed is essential for applications requiring immediate responses. Techniques such as model acceleration, parallel processing, and efficient tokenization strategies can contribute to achieving real-time performance.

\noindent\textbf{Multimodal Expansion.} Extending VideoLLaMA3 to incorporate additional modalities like audio, speech, and sensor data can create a more holistic understanding of multimodal inputs. Research into unified architectures that seamlessly integrate multiple data types will be pivotal in achieving comprehensive multimodal intelligence.

\noindent\textbf{Advanced Post-Training Techniques.} Implementing more sophisticated post-training methodologies, such as scaling RL techniques for MLLMs, can further refine VideoLLaMA3's performance. RLHF and other RL-based approaches can be employed to better align the model's outputs with human preferences and task-specific requirements. Scaling these RL techniques to accommodate the complexities of multimodal data will enhance the model's ability to generate more accurate, contextually appropriate, and user-aligned responses, thereby advancing its overall multimodal intelligence.

In summary, while VideoLLaMA3 represents a significant step forward in multimodal AI, addressing its current limitations through targeted research and development will pave the way for even more powerful and versatile models in the future.